\documentclass[lettersize,journal]{IEEEtran}
\usepackage{amsmath,amsfonts}
\usepackage{amsthm}
\newtheorem{definition}{Definition}
\usepackage{stmaryrd}
\usepackage{algorithmic}
\usepackage{algorithm}
\usepackage{array}
\usepackage[caption=false,font=normalsize,labelfont=sf,textfont=sf]{subfig}
\usepackage{textcomp}
\usepackage{stfloats}
\usepackage{url}
\usepackage{verbatim}
\usepackage{graphicx}
\usepackage{amssymb}

\usepackage{multirow}

\usepackage{xcolor}
\usepackage{soul}
\newif\ifshowhl
\showhlfalse

\sethlcolor{red!20}

\soulregister\cite7
\soulregister\ref7
\soulregister\eqref7
\soulregister\textit7
\soulregister\textbf7

\newcommand{\rev}[1]{\ifshowhl\hl{#1}\else#1\fi}
\newcommand{\Romannum}[1]{\uppercase\expandafter{\romannumeral #1\relax}}

\usepackage{cite}
\begin{document}

\title{Two-dimensional Taxonomy for $n$-ary Knowledge Representation Learning Methods}

\author{Xiaohua Lu, Liubov Tupikina, and Mehwish Alam%
\thanks{Xiaohua Lu (Corresponding author) is with the University of Rouen Normandy, Rouen, France (e-mail: xh.lu0706@gmail.com).}%
\thanks{Liubov Tupikina is with Nokia Bell Labs, Paris, France.}%
\thanks{Mehwish Alam is with Télécom Paris, Institut Polytechnique de Paris, Paris, France (e-mail: mehwish.alam@telecom-paris.fr).}%
\thanks{Manuscript received June 27, 2025; revised June 8, 2026; accepted August 29, 2026.}}

\markboth{IEEE Transactions on Knowledge and Data Engineering}%
{Lu \MakeLowercase{\textit{et al.}}: Two-dimensional Taxonomy for $n$-ary Knowledge Representation Learning Methods}


% First-page notices for the author's accepted manuscript on arXiv.
\AddToHookNext{shipout/foreground}{%
  \put(0pt,-42pt){\makebox[\paperwidth]{\parbox[t]{\textwidth}{\centering\normalfont\fontsize{8}{9}\selectfont
    This is the author's accepted manuscript of an article accepted for publication in IEEE Transactions on Knowledge and Data Engineering.}}}%
  \put(0pt,\dimexpr-\paperheight+18pt\relax){\makebox[\paperwidth]{\parbox[b]{\textwidth}{\normalfont\fontsize{7}{8.5}\selectfont
    \copyright\ 2026 IEEE. Personal use of this material is permitted. Permission from IEEE must be obtained for all other uses, in any current or future media, including reprinting/republishing this material for advertising or promotional purposes, creating new collective works, for resale or redistribution to servers or lists, or reuse of any copyrighted component of this work in other works.}}}%
}

\maketitle

%%%%%%%%%%%%%%%%%%%%%%%%%%%%%%%%%%%%%%%%%%%%%%%%%%%%%%%%%%%%%%%%%%%%%%%%
\begin{abstract}

Real-world knowledge can take various forms, including structured, semi-structured, and unstructured data. Among these, Knowledge Graphs (KGs) are structured representations that integrate heterogeneous data sources into structured representations. However, KGs typically reduce complex \(n\)-ary relations to simple triples, thereby losing higher-order relational details.
In contrast, hypergraphs naturally represent \(n\)-ary relations with hyperedges that directly connect multiple entities. 
Recent advances have led to many hypergraph representation learning methods; however, they often overlook entity roles in hyperedges, limiting fine-grained semantic modelling.
% Recent advances have led to many methods which learn representations over hypergraphs, however, it often overlooks entity roles in hyperedges, limiting fine-grained semantic modelling.
To address these issues, Knowledge Hypergraphs (KHGs) and Hyper-relational Knowledge Graphs (HKGs) combine the advantages of KGs and hypergraphs to better capture complex relational structures and role-specific semantics of real-world knowledge.
This survey provides a \rev{structured review} of representation learning methods for \(n\)-ary relational data, with a \rev{primary focus on static} KHGs and HKGs.
We propose a two-dimensional taxonomy: the first dimension categorises models based on their methodology, including translation-based models, tensor factorisation-based models, deep neural network-based models, logic rule-based models, and hyperedge expansion-based models. The second dimension classifies models according to their awareness of entity roles and positions in \(n\)-ary relations, dividing them into position-aware, role-aware, and aware-less approaches.
\rev{Finally, we summarise benchmark datasets, training settings, negative sampling strategies, and benchmark-style performance comparisons, and outline open challenges to inspire future research.}

\end{abstract}


%%%%%%%%%%%%%%%%%%%%%%%%%%%%%%%%%%%%%%%%%%%%%%%%%%%%%%%%%%%%%%%%%%%%%%%%
\section{Introduction}

\begin{figure*}[!t]
    \centering
    \includegraphics[width=0.85 \textwidth]{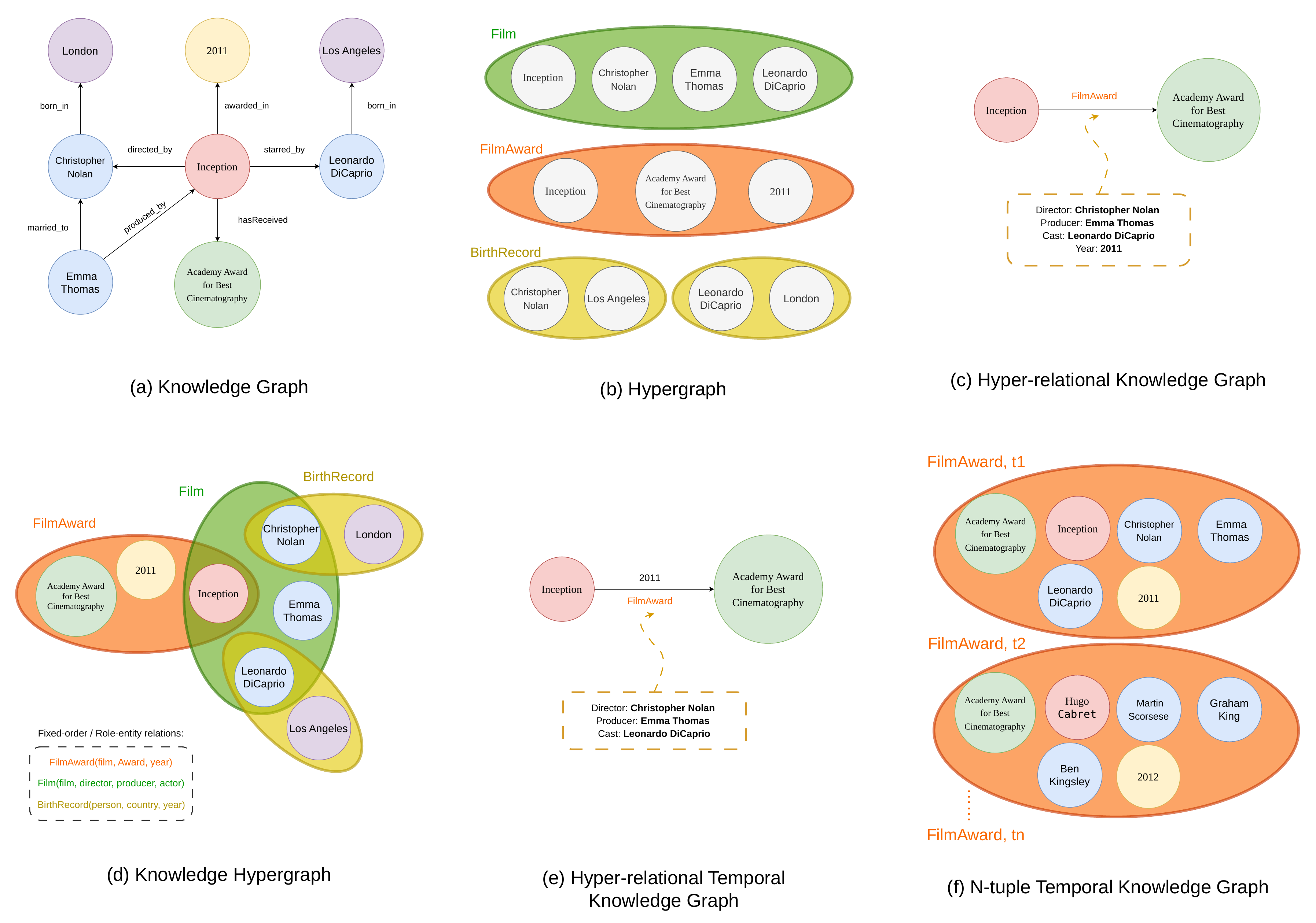}
    \caption{Illustrative examples of graph-based knowledge representation formalisms. (a) Knowledge Graph (KG), (b) Hypergraph (HG), (c) Hyper-relational Knowledge Graph (HKG), (d) Knowledge Hypergraph (KHG), (e) Hyper-relational Temporal Knowledge Graph (HTKG), and (f) N-tuple Temporal Knowledge Graph (N-TKG).}

    % \caption{Comparison of KG and HG. 
    % (a) A standard KG models facts about the film Inception using binary relations. (b) A HG represents co-authorship in a research network, capturing higher-order group interactions via hyperedges.}
    \label{fig:introFig1}
\end{figure*}

\IEEEPARstart{W}{ITH} the increasing interest in KGs~\cite{KGsurvey2021} in areas such as education \cite{aliyu2020}, scientific research \cite{chi2018knowledge}, social sciences~\cite{ChenSA22}, healthcare \cite{gong2021smr}, and finance \cite{cheng2022financial}, the design of KGs capable of supporting complex reasoning and semantic modelling has become a key research topic \cite{KGsurvey2023}. 
% KGs represent knowledge in a structured triple, such as \(\langle\)head entity, relation, tail entity\(\rangle \), the semantic descriptions of entities can also be considered as literals \cite{alam2024neurosymbolic}. 
KGs usually represent knowledge as structured triples, such as (head entity, relation, tail entity). In addition to entities and relations, KGs may also include literals as auxiliary information, such as textual descriptions, numerical values, and visual features \cite{alam2024neurosymbolic}.
In recent years, a number of large-scale open-source KGs have been designed, including WordNet \cite{2004wordnet}, DBpedia \cite{2007dbpedia}, YAGO \cite{SuchanekABCPS24}, Freebase \cite{Freebase2008}, and Wikidata \cite{wikidata2014}. Figure~\ref{fig:introFig1}(a) shows an example of a KG related to the film \textit{Inception}.

To enable downstream tasks such as KG completion \cite{KGcompletion2020,KGcompletionb2020}, question answering \cite{MHPGM2018, PCQA2019, EmbedKGQA2020}, recommender systems \cite{Survey2024}, and reasoning \cite{KGreasonb2019, KGreason2019}, KG representation learning (KGRL), also called KG embedding, has emerged as a key technique. It aims to encode entities and relations into low-dimensional vectors while preserving their semantic and structural properties \cite{KGsurvey2023}. However, KGRL models can only handle binary relations, which are insufficient to capture the multi-entity interactions commonly observed in the real world. 

In practice, many facts involve more than two entities in a relation, forming \(n\)-ary interactions, also called higher-order relations \cite{HGsurvey2024}. To overcome this limitation, hypergraphs (HGs) have been introduced as a natural extension of traditional graphs. Unlike standard graphs where edges connect only two nodes, HGs allow each hyperedge to connect multiple nodes simultaneously, allowing direct representation of \(n\)-ary interactions. 
As shown in Figure \ref{fig:introFig1}(b), for example, the hyperedge \emph{Film} groups together with the movie \textit{Inception} and the people involved in it (director \textit{Christopher Nolan} and producer \textit{Emma Thomas} and cast \textit{Leonardo DiCaprio}). Each coloured ellipse therefore represents one $n$-ary relation involving more than two entities, rather than a collection of pairwise edges. Moreover, all entities inside a given hyperedge are drawn as grey circles, indicating that HGs only specify which entities participate together, but do not distinguish their semantic roles or argument types. 
Beyond this toy example, many real-world group interactions can be naturally modelled as HGs, such as: a co-authored paper typically involves multiple authors \cite{2019hplapgcn,NHNE2020,DHGNN2019,dhe2019}, and a community in a social network may contain many members \cite{soial2022,social2019}.
Accordingly, HG representation learning (HGRL) has become an important approach to derive meaningful embeddings for nodes or hyperedges.
Despite recent advances, achieving scalable and efficient structural integration remains a key challenge due to high computational cost and complex model design \cite{2021netvec,shun2020,hgchallenge2021}.

To address this gap, recent work has introduced the concept of \textit{$n$-ary KGs (NKGs)} \cite{2025survey}, aiming to combine the expressive power of HGs with the semantic richness of KGs. Two predominant forms of NKG structures have emerged: KHGs \cite{fatemi2019} and HKGs \cite{neuinfer2020}. 
In Figure~\ref{fig:introFig1}(c), the film-award fact is represented as an HKG: the binary triple (\textit{Inception}, \textit{FilmAward}, \textit{Academy Award for Best Cinematography}) is enriched with qualifiers that specify the director, producer, cast, and award year as qualifier pairs attached to the edge. 
In Figure~\ref{fig:introFig1}(d), the same information is modelled as a KHG, where each coloured hyperedge (\textit{Film}, \textit{FilmAward}, \textit{BirthRecord}) links all participating entities in a fixed-order, role-typed tuple, e.g.,  \textit{Film}(film, director, producer, actor) or \textit{BirthRecord}(person, country, year), making the roles of each entity explicit and contrasting with the role-agnostic hyperedges in Figure~\ref{fig:introFig1}(b).

KHGs use directed, labelled hyperedges to encode \(n\)-ary facts, while HKGs enrich the primary triples with qualifier pairs. These representations mitigate the drawbacks of star-to-clique transformation or reification techniques (see Supplementary Material, Section~\Romannum{1}.A), that convert \(n\)-ary facts into multiple binary forms. These transformations often suffer from limitations, including loss of higher-order structures, increased knowledge sparsity, and over-parameterisation problems \cite{fatemi2019}. Therefore, the direct modelling of \(n\)-ary facts via KHGs and HKGs has received increasing attention in recent years \cite{fatemi2019, neuinfer2020,Survey2024}. 

\rev{For completeness, Figure~\ref{fig:introFig1} also illustrates two temporal extensions of $n$-ary knowledge representations. Although this survey primarily focuses on static $n$-ary relational representation learning, temporal information is relevant in many KGs and has recently started to be studied in $n$-ary and hyper-relational settings. In Figure~\ref{fig:introFig1}(e), the hyper-relational temporal KG (HTKG) attaches time to the main triple while representing additional information as qualifier pairs. In Figure~\ref{fig:introFig1}(f), the $n$-tuple temporal KG (N-TKG) encodes time as an additional argument within each $n$-ary fact. Only a few recent models, such as HypeTKG~\cite{hyperTKG2023} and NE-Net~\cite{NEnet2023}, have explored these temporal extensions.
Therefore, temporal $n$-ary knowledge representation is not the main focus of this survey. We mention these temporal extensions only briefly to clarify related formalisms and to delimit the scope of the paper.
% Therefore, we discuss them only briefly where relevant, and leave a comprehensive survey of temporal $n$-ary knowledge representation as future work.
}

As these new structures gain traction, a key challenge emerges: how to learn effective representations from them. Existing methods can be broadly divided into two groups: \textit{KHG Representation Learning} (KHGRL) that directly extend traditional KGRL or HGRL models to hyperedges, such as m-TransH \cite{mtransH2016}, m-CP \cite{fatemi2019}, G-MPNN \cite{GMPNN2020}, and H\textsuperscript{2}GNN \cite{H2GNN2024}, and \textit{HKG Representation Learning} (HKGRL) that encode qualifier pairs to enrich the original KG representation, such as NaLP \cite{NaLP2021}, NeuInfer \cite{neuinfer2020}, and STARE \cite{stare2020}. 
\rev{These methods have developed rapidly in recent years, but they remain scattered across different modelling assumptions, data structures, and evaluation settings.}

\rev{
This growing body of work has motivated the need for survey-level synthesis.
Although a recent survey has provided an important overview of link prediction in NKGs~\cite{2025survey}, existing surveys remain mainly centred on technical architectures or specific downstream tasks.
They provide limited analysis of how models differ in semantic awareness, qualifier modelling, training strategies, and broader design principles for $n$-ary relational representation learning.
To complement these works, our survey proposes a dual-axis taxonomy that disentangles technical modelling principles from semantic awareness, and further provides a structured synthesis of benchmark datasets, training settings, negative sampling strategies, model comparison results, and open challenges.
}

\paragraph{\textbf{Contributions}} Our contributions can be summarised as follows:
\begin{itemize}
  \item We unify the terminology and formalism for static \(n\)-ary relational data, including KHGs and HKGs, and clarify the corresponding link prediction tasks. 
  % We also briefly discuss recent temporal extensions and their associated prediction settings as an emerging research direction.

  \item We propose a novel two-dimensional taxonomy for \(n\)-ary relational representation learning models. \rev{Unlike purely technique-oriented classifications, our taxonomy jointly considers technical modelling principles and semantic awareness, allowing a more fine-grained comparison of how models capture entity positions, semantic roles, and qualifier structures. 
  We cover over 50 representative models published by the end of 2025.
  }

  \item \rev{We provide a systematic cross-model analysis of existing methods, highlighting their modelling assumptions, semantic expressiveness, computational characteristics, and limitations. Based on this analysis, we identify the key characteristics of effective $n$-ary relational representation learning models and formulate design principles for future model development.}

  \item We systematically summarise existing benchmark datasets, training settings, and negative sampling strategies for \(n\)-ary relational data. \rev{We also provide a benchmark-style comparison of reported model performance where comparable results are available, while discussing the difficulty of fully unified comparison due to heterogeneous datasets, splits, and evaluation protocols.}

  \item Finally, we outline future directions for \(n\)-ary relational data modelling, including temporal extensions, multimodal extension, inductive learning, end-to-end architectures, and scalable and interpretable learning. We also highlight the need for more standardised benchmark suites and evaluation protocols for KHG and HKG models.
\end{itemize}

\paragraph{\textbf{Article organisation}} 
\rev{
Section~\ref{sec:preliminaries} introduces the main notations, structural definitions, and link prediction tasks for $n$-ary relational data. 
Section~\ref{sec:related-work} reviews existing surveys on KGRL, HGRL, and NKGs, and compares their scopes and taxonomies. 
Section~\ref{sec:taxonomy} presents our two-dimensional taxonomy, model classification, temporal extensions, complexity analysis, and modelling guidelines. 
Section~\ref{sec:ds} summarises benchmark datasets, negative sampling strategies, and reported KHG/HKG model performances. 
Finally, Section~\ref{sec:conclusion} concludes the paper and discusses open challenges.
}

% \paragraph{\textbf{Article organisation}} Section~\ref{sec:preliminaries} introduces the notations and definitions for modelling \(n\)-ary knowledge data and models. Section~\ref{sec:related-work} reviews existing surveys in the fields of NKGs, highlights their classification strategies, and identifies the lack of a unified perspective on \(n\)-ary knowledge representation.
% % Section~\ref{sec:related} provides a quick review of prior work on KGRL and HGRL methods. 
% Section~\ref{sec:taxonomy} presents our two-dimensional taxonomy and detailed classification of existing models, gives a complexity analysis of models and summarises a guideline for good $n$-ary representation models. Section~\ref{sec:ds} summarises some available datasets and discusses negative sampling strategies. Finally, Section~\ref{sec:conclusion} concludes the paper and outlines future research opportunities.

%%%%%%%%%%%%%%%%%%%%%%%%%%%%%%%%%%%%%%%%%%%%%%%%%%%%%%%%%%%%%%%%%%%%%%%%
\section{Preliminaries} \label{sec:preliminaries}
This section introduces the fundamental concepts and notations used throughout this survey. We formally define the KGs \cite{alam2024neurosymbolic,KGsurvey2021}, HGs \cite{HGsurvey2023}, KHGs \cite{PosKHG2023}, and HKGs \cite{stare2020}. We also introduce $n$-TKGs \cite{NEnet2023} and HTKGs \cite{hyperTKG2023} as two temporal extensions of KHGs and HKGs, respectively. These structures enrich static knowledge representations by incorporating temporal attributes. 
We further define the link prediction tasks, also called KG completion task \cite{QUAD2023}, tailored to both KHG \cite{PosKHG2023} and HKG settings \cite{GRAN2021}. 
Finally, to describe the expressiveness of link prediction models over \(n\)-ary relational data, we introduce the notion of full expressiveness \cite{fatemi2019}. 

% A fully expressive model can capture and represent different relation patterns through embeddings of \(n\)-ary facts. In contrast, models lacking full expressiveness can only capture limited knowledge with strong prior assumptions \cite{GETD2020}, potentially leading to unwarranted inferences \cite{fatemi2019}.

\subsection{Structural Definitions}
% Hypergraph
\begin{definition}[\textbf{Hypergraph}]
\normalfont
A HG is a generalisation of a conventional graph defined as \(\mathcal{H} = (\mathcal{V}, \mathcal{E})\), where \(\mathcal{V}\) is a finite set of vertices (or nodes) and \(\mathcal{E}\) is a finite set of hyperedges. Each hyperedge is a non-empty subset of  \(\mathcal{V}\), containing arbitrary vertices. In addition, the HG is associated with an incidence matrix \(\mathcal{I} \in \{0,1\}^{|\mathcal{V}|\times|\mathcal{E}|}\) defined by 
% \[
% \mathcal{I}(v,e)=\llbracket v\in e\rrbracket
% \]
\begin{equation}
\mathcal{I}(v,e)=\llbracket v\in e\rrbracket,
\end{equation}
where \(v \in \mathcal{V}\), \(e \in \mathcal{E}\) and \(\llbracket p\rrbracket\) is the indicator function which equals to 1 if the predicate \(p\) is true and 0 otherwise.
This definition follows the formulation in~\cite{HGdef2006}, which builds on the classical notion introduced by Berge in 1984 \cite{1984HGs}.
\end{definition}

% Knowledge Graph
\begin{definition}[\textbf{Knowledge Graph}]
\normalfont
A KG is defined as a directed labelled graph \(\mathcal{G} = (\mathcal{E}, \mathcal{R}, \mathcal{L}, \mathcal{F})\), where \(\mathcal{E}\) is a finite set of entities representing real or abstract concepts; \(\mathcal{R}\) is a finite set of relations between entities; \(\mathcal{L}\) is a finite set of literals, such as numerical values, texts, or images; and 
\begin{equation}
\mathcal{F} \subseteq \mathcal{E} \times \mathcal{R} \times (\mathcal{E} \cup \mathcal{L})
\end{equation}
is a finite set of facts. Each fact \(f \in \mathcal{F}\) is expressed as a triple \((h, r, t)\) or as \((h, r, l)\), where \(h, t \in \mathcal{E}\), \(r \in \mathcal{R}\), and \(l \in \mathcal{L}\).
This definition extends the canonical triple-based formulation \cite{transE2013} by including \textit{literals} to enable richer semantic representations as discussed in \cite{KGliteral2019, alam2024neurosymbolic}.

% Although the terms "knowledge base" and "knowledge graph" are sometimes used interchangeably in the literature, KGs tend to emphasize graph-oriented organization and explicit semantic relationships  \cite{KGchallenges2023}.
\end{definition}

\begin{definition}[\textbf{Knowledge Hypergraph}]
\normalfont
A KHG is defined as \(\mathcal{K} = (\mathcal{E}, \mathcal{R}, \mathcal{T}_O)\), where \(\mathcal{E}\) is a finite set of entities, \(\mathcal{R}\) is a finite set of relations, and \(\mathcal{T}_O\) is a finite set of observed relational facts.
Each relation \(r \in \mathcal{R}\) is associated with an arity \(\alpha \geq 2\), and each tuple \(t \in \mathcal{T}_O\) takes the form:

\begin{equation}\label{eq:khg-def1}
t = r(e_1, e_2, \dots, e_\alpha),
\end{equation}
where \(e_i \in \mathcal{E}\) for all \(i\). The positions of entities \(e_i\) in the tuple are fixed and ordered, implicitly encoding their semantic roles. 

Alternatively, a tuple can be associated with a set $M(r)$ of ordered roles $\{\rho_r^1, \rho_r^2, \dots, \rho_r^\alpha\}$:
\begin{equation}\label{eq:khg-def2}
t = r(\rho_r^1 : e_1, \rho_r^2 : e_2, \dots, \rho_r^\alpha : e_\alpha),
\end{equation}
where \(\rho_r^i\) denotes the explicit role of \(e_i\) in relation \(r\).\\
% The definition in Eq.~\ref{eq:khg-def1} synthesises the formalisation proposed by Fatemi et al.~\cite{fatemi2019}, while Eq.~\ref{eq:khg-def2} provides an equivalent role-annotated notation for making relation-specific argument roles explicit. This notation is related to early formulations of n-ary relational facts, but it does not imply that all corresponding embedding models explicitly encode roles.
The definition in Eq. \ref{eq:khg-def1} synthesises the formalisation proposed by Fatemi et al. \cite{fatemi2019}, while the role-aware formulation in Eq. \ref{eq:khg-def2} follows the modelling assumptions of RAM~\cite{RAM2021}. These two works represent some of the earliest KHG-based models.
\end{definition}

% Hyper-relational Knowledge Graph

\begin{definition}[\textbf{Hyper-relational Knowledge Graph}]
\label{def:hkg}
\normalfont
A HKG is defined as \(\mathcal{HKG} = (\mathcal{E}, \mathcal{R}, \mathcal{H})\), where \(\mathcal{E}\) is a finite set of entities, \(\mathcal{R}\) is a finite set of relations, and \(\mathcal{H}\) is a list of hyper-relational facts. Each hyper-relational fact is represented as:

\begin{equation} \label{eq:hkg}
f = (s, r, o, \mathcal{Q}), with \ \mathcal{Q} = \{(a_1, e_1), \dots, (a_k, e_k)\}
\end{equation}

where \((s, r, o) \in \mathcal{E} \times \mathcal{R} \times \mathcal{E}\) denotes the primary triple. \(\mathcal{Q} \in \mathcal{P}(\mathcal{R} \times \mathcal{E})\) is a set of qualifier pairs, with \(a_i \in \mathcal{R}\) and \(e_i \in \mathcal{E}\). The set \(\mathcal{P}(\cdot)\) denotes the power set. 
The definition in Eq. \ref{eq:hkg} follows the definition introduced by STARE~\cite{stare2020}.
\end{definition}

% TKG

\begin{definition}[\textbf{Temporal Knowledge Graph}]
\label{def:tkg}
\normalfont
A TKG is a temporal extension of a static KG, where relational facts are associated with temporal information. Two common formalisms are widely adopted in the literature:
\begin{itemize}
    \item \textbf{Event-based form}: A TKG is represented as a set of timestamped facts 
    \begin{equation}
    \mathcal{T} = \{(s, r, o, t)\},    
    \end{equation}
    where \(t\) denotes the timestamp when the fact holds.
    \item \textbf{Snapshot-based form}: A TKG is viewed as a sequence of KG snapshots
     \begin{equation}
    \mathcal{T} = \{\mathcal{G}_1, \mathcal{G}_2, \dots, \mathcal{G}_T\},    
    \end{equation}
    where each \( \mathcal{G}_t \) corresponds to a static KG at time \( t \).
\end{itemize}

Pioneering works introducing the event-based and snapshot-based formulations include \cite{TKGdef12018} and \cite{TKGdef22020}, respectively. 
\end{definition}

% HTKG
\begin{definition}[\textbf{Hyper-relational Temporal Knowledge Graph}]
\label{def:h-tkg}
\normalfont
% check HypeTKG
Let \(\mathcal{E}\), \(\mathcal{R}\), and \(\mathcal{T}\) denote the finite sets of entities, relations, and timestamps, respectively. A HTKG is defined as a set of time-stamped hyper-relational facts. Each fact is represented as: 
\begin{equation}
    f = ((s, r, o, \tau), \{(a_i,v_i)\}_{i=1}^{n}),
\end{equation}
where \((s, r, o, \tau)\) is the primary quadruple, and $\tau \in T$ denotes the timestamp. Each qualifier \((a_i, v_i)\) consists of a qualifier relation \(a_i \in \mathcal{R}\) and a qualifier entity \(v_i \in \mathcal{E}\). The integer \(n\) denotes the number of qualifiers. 

This formulation is based on the HypeTKG model~\cite{hyperTKG2023}, one of the earliest models explicitly introducing hyper-relational temporal KGs.
\end{definition}

% $n$-TKG
\begin{definition}[\textbf{N-Tuple Temporal Knowledge Graph}]
\normalfont
\label{def:$n$-tkg}
% check Ne-NET
An N-TKG is a sequence of timestamped KGs \(\mathcal{G} = \{\mathcal{G}_1, \mathcal{G}_2, \dots, \mathcal{G}_t\}\). Each \(\mathcal{G}_t = (\mathcal{V}_{\text{pred}}, \mathcal{V}_{\text{ent}},\mathcal{V}_{\rho}, \mathcal{F}_t)\), where \(\mathcal{V}_{\text{pred}}\) is a finite set of predicates, \(\mathcal{V}_{\text{ent}}\) is a finite set of entities, \(\mathcal{V}_{\rho}\) is a finite set of roles, and \(\mathcal{F}_t\) is a set of facts at time \(t\). A fact \(f \in \mathcal{F}_t\) is expressed as:
\begin{equation}
    f = (\text{pred}, \rho_1{:}e_1, \dots, \rho_n{:}e_n, t),    
\end{equation}
where \(\text{pred}\) is a predicate, \(e_i\) is an entity, \(\rho_i\) is its role, and \(t\) is the timestamp. 
This definition follows the formulation proposed in the NE-Net model \cite{NEnet2023}.
\end{definition}

% % Representation Learning
% \begin{definition}[\textbf{(Knowledge Graph) Representation Learning}]
%     % check https://yiyibooks.cn/arxiv/2402.19414v2/index.html  sec3.2.5
% \end{definition}

\subsection{Task Formulations and Model Property}
% Link Prediction Task
\begin{definition}[\textbf{Link Prediction for KHG}]
\normalfont
In a KHG, an \(n\)-ary fact is represented as an \(n\)-ary tuple:
\(t = r(\rho^1_r: e_1, \rho^2_r: e_2, \dots, \rho^\alpha_r: e_\alpha)\),
where \(r\) is a relation, \(e_i\) are entities, and \(\rho^i_r\) denotes the role of \(e_i\) in relation \(r\).
Let \(\mathcal{T}_O \subseteq \mathcal{T}_T \subseteq \mathcal{T}\) denote the observed tuples, the ground-truth tuples, and all possible \(n\)-ary tuples, respectively. The set of hidden tuples is defined as \(\mathcal{T}_H = \mathcal{T} \setminus \mathcal{T}_O\). The goal of link prediction (LP) in KHGs is to predict the labels of hidden tuples \(\mathcal{T}_H\), given the observed tuples \(\mathcal{T}_O\), i.e., to identify which unobserved tuples are likely to be true.
For example, given the incomplete \(n\)-ary fact: \(r(e_1, e_2, ?, e_4)\), the goal is to predict the missing object entity \(e_3\).
This formulation is consistent with early definitions adopted for $n$-ary facts in KHGs~\cite{fatemi2019}.

\end{definition}

% Link Prediction Task

\begin{definition}[\textbf{Link Prediction for HKG}]
\normalfont
A hyper-relational fact typically consists of a primary triple \((s, r, o)\) together with a set of qualifier pairs \(\{(a_i : v_i)\}_{i=1}^m\), where \(s, o, v_i\) are entities and \(r, a_i\) are relations.  
LP in HKGs refers to the task of inferring missing elements from an incomplete hyper-relational fact. 
This may involve predicting a missing part of the primary triple, such as the subject $s$, relation $r$, or object $o$ in traditional triple completion tasks. For example, given an incomplete fact $((s, r, ?), \{(a_i : v_i)\}_{i=1}^m)$, the goal may be to predict the object $o$ of the primary triple. 

Alternatively, it may involve predicting missing components within the qualifiers, such as $a_i$ or its corresponding value $v_i$, depending on the model design and application context. In this setting, given a fact such as $((s,r,o), \{(a_1:v_1), \ldots, (a_k:?), \ldots\})$, the goal is to predict the missing qualifier value $v_k$.
% given a fact such as $((s, r, o), \{(a_1 : v_1), ..., (? : v_k), ...\})$, the goal is to predict the missing $a_k$ in the qualifier set.

The definition of primary triple completion is from STARE \cite{stare2020}, while the formulation of qualifier completion first appears in Hy-Transformer \cite{HyTransformer2021}.
\end{definition}

\begin{definition}[\textbf{Transductive and Inductive LP on HKGs}]
\normalfont
Consider an HKG \(G = (E, R, H)\), where \(E\) is a set of entities, \(R\) is a set of relations, and \(H\) is a set of hyper-relational facts.
% In a HKG $G = (E, R, H)$ , where $E$ is a set of entities, $R$ is a set of relations, and $H$ is a set of hyper-relational facts. 

In \textbf{transductive LP}, the graph is split into three pairwise disjoint sets: training facts  $H_\mathrm{tr}$, validation facts $H_\mathrm{val}$, and test facts $H_\mathrm{tst}$, i.e., $H = H_\mathrm{tr} \cup H_\mathrm{val} \cup H_\mathrm{tst}$. 
The goal is to predict a missing entity in a test fact $h \in H_\mathrm{tst}$, assuming all entities and relations were seen during training.

In \textbf{inductive LP}, we are given a training HKG $G = (E, R, H)$ and a separate inference HKG $G' = (E', R', H')$, where $E' \not\subseteq E$ and/or $R' \not\subseteq R$. The inference graph $G'$ is similarly split into observed facts $H'_\mathrm{fct}$, validation facts $H'_\mathrm{val}$, and test facts $H'_\mathrm{tst}$. 
% The task is to predict a missing entity in a test fact $h \in H'_{tst}$, using only the observed facts in the inference graph, without retraining on $G'$.
\rev{The task is to predict a missing entity in a test fact $h \in H'_\mathrm{tst}$, using only the observed facts in the inference graph, without retraining on $G'$.}
% The task is to predict a missing entity in a test fact $h \in H'_\mathrm{tst}$, using only the structure of $H'_\mathrm{tst}$, without retraining on $G'$.
These formulations follow one of the earliest papers that discuss the transductive and inductive LP on HKGs \cite{maypl2025}.
\end{definition}

% Fully Expressive Models
\begin{definition}[\textbf{Property: Full Expressiveness}] 
\label{def:fullexp}
\normalfont
A link prediction model is \emph{fully expressive} if, for any ground truth over all entities and relations, there exist entity and relation embeddings that accurately distinguish true \(n\)-ary relational facts from false ones. \\
This notion was first introduced by Fatemi et al. \cite{fatemi2019} and later restated in a more general form in the ReAIE model \cite{ReAIE2023}.
% \rev{An illustrative example of full expressiveness is provided in }supplementary material \Romannum{2}.

\paragraph{Illustrative example} Consider a minimal $n$-ary relational world:

\begin{itemize}
    \item Entities: Schwarzenegger, T-800, Terminator2, California;
    \item Relations:
    \begin{itemize}
        \item PlayCharacterIn (Actor, Character, Movie);
        \item Governed (Person, Region).
    \end{itemize}
\end{itemize}
\begin{itemize}
    \item True facts:\\
    (1) \textbf{PlayCharacterIn}(Actor: Schwarzenegger, Character: T-800, Movie: Terminator2);\\
    (2) \textbf{Governed}(Person: Schwarzenegger, Region: California).
    \item False facts:\\
    (3) \textbf{PlayCharacterIn}(Actor: Schwarzenegger, Character: Schwarzenegger, Movie: Terminator2);\\
    (4) \textbf{Governed}(Person: T-800, Region: California).
\end{itemize}
A model lacking full expressiveness may fail to separate different semantic roles associated with the entity. 
For instance, if \texttt{Schwarzenegger} and \texttt{T-800} are assigned one embedding regardless of their role (e.g., Actor vs. Character), the model may incorrectly score the false fact (3) as likely true.
In contrast, a fully expressive model assigns distinct embeddings for each entity-role-position combination, allowing it to correctly identify true facts.

\end{definition}

\section{Related Works} \label{sec:related-work}

\begin{table*}[t]
\centering
\caption{\rev{Comparison of existing surveys related to $n$-ary relational representation learning}.}
\label{tab:survey_comparison}
\scriptsize
\setlength{\tabcolsep}{3pt}
\renewcommand{\arraystretch}{1.15}
\begin{tabular}{|p{1.5cm}|p{1.8cm}|p{1.5cm}|p{3.0cm}|p{1.cm}|p{1.3cm}|p{2.6cm}|p{3.2cm}|}
\hline
\textbf{Survey} 
& \textbf{Main scope} 
& \textbf{Main tasks} 
& \textbf{\(n\)-ary KG formalism coverage} 
& \textbf{Temporal coverage} 
& \textbf{Qualifier prediction} 
& \textbf{Dataset / benchmark synthesis} 
& \textbf{Taxonomy / analytical dimensions} \\
\hline

KGRL~\cite{KGsurvey2017,KGsurvey2021,KGsurvey2023,KGsurvey2024}
& Binary KG representation learning
& KG completion, reasoning
& --
& --
& --
& --
& Technical families, learning paradigms, scoring functions, or representation spaces \\
\hline

HGRL~\cite{HGsurvey2020,HGsurvey2022,HGsurvey2023,HGsurvey2024}
& Hypergraph representation learning
& Node/link prediction
& HG structure only; no knowledge-centric roles or qualifiers
& --
& --
& --
& Task-, theory-, or technique-based taxonomies \\
\hline

NKG Survey ~\cite{2025survey}
& \(n\)-ary KG representation learning
& Link prediction
& Hyperedge, role-value pair, and hyper-relational formalizations
& Brief
& Not a main focus
& 3 datasets
& Technique-based taxonomy + application / special scenarios \\
\hline

\textbf{Our survey}
& \(n\)-ary relational representation learning
& \textbf{Link prediction; qualifier prediction}
& KHGs and HKGs, with semantic-awareness analysis
& Brief
& \textbf{Discussed explicitly}
& \textbf{10+ datasets}; training, sampling, \textbf{benchmark-performance synthesis}
& \textbf{Two-dimensional taxonomy}: technique + semantic awareness \\
\hline

\multicolumn{8}{|p{18.0cm}|}{\footnotesize \textit{Note}: ``--'' indicates that the corresponding aspect is not considered as a comparison criterion, rather than being necessarily absent from the cited surveys.} \\
\hline
\end{tabular}
\end{table*}

In recent years, a number of surveys have systematised the development of KG representation learning (KGRL) and Hypergraph Representation Learning (HGRL). Early KGRL surveys mainly organise embedding models according to technical families, such as translation-based, semantic matching-based, and auxiliary information-enhanced methods~\cite{KGsurvey2017}. 
Subsequent surveys broaden this view by considering learning paradigms, scoring functions, encoding models, and the use of auxiliary information~\cite{KGsurvey2021,KGsurvey2023}. 
More recently, Cao et al.~\cite{KGsurvey2024} revisited KGRL from the perspective of mathematical representation spaces, including algebraic, geometric, and analytic spaces.
Despite this evolution, these surveys mainly focus on binary KGRL and assume triple-based facts, which limits their applicability to \(n\)-ary tuples and qualifier-based hyper-relational facts.

In parallel, HGRL surveys have developed along a different line. 
Gao et al.~\cite{HGsurvey2020} provide a task-oriented overview, covering transductive and inductive hypergraph learning, structure updating, and multimodal hypergraph learning. 
Other surveys organise HGRL methods according to theoretical foundations or technical mechanisms, such as matrix decomposition, random walks, spectral methods, proximity-preserving methods, and deep neural networks~\cite{HGsurvey2022,HGsurvey2023}. 
More recently, Tian et al.~\cite{HGsurvey2024} discuss higher-order network representations, including network motifs, simplicial complexes, and HGs, and classify related methods from a broader higher-order modelling perspective. 
These surveys successfully characterise higher-order structural learning, but they generally focus on combinatorial hypergraph structures rather than knowledge-centric modelling. 
As a result, they do not fully address typed relations, qualifiers, entity roles, or role-specific semantics, which are central to \(n\)-ary knowledge representation. 
For a more detailed technical background on KG and HG models, we refer the reader to Supplementary Material, Section~\Romannum{1}.

The gap between KGRL and HGRL surveys has motivated recent work on NKGs, including KHGs and HKGs, which aim to jointly model higher-arity interactions and richer relational semantics.
\rev{Wei et al.~\cite{2025survey} present the first dedicated survey on link prediction in NKGs.
They classify nearly forty methods into spatial mapping-based, tensor decomposition-based, and deep neural network-based methods, and compare reported performance on three standard benchmarks: JF17K~\cite{mtransH2016}, WikiPeople~\cite{NaLP2021}, and WD50K~\cite{stare2020}.
This work represents an important step towards structuring NKG link prediction research.
However, its scope is mainly centred on link prediction, and its taxonomy remains primarily technique-oriented.

For example, within tensor-based methods, m-DistMult~\cite{fatemi2019} mainly extends traditional tensor factorisation to $n$-ary structures without explicit semantic awareness, whereas HypE~\cite{fatemi2019} models positional information, and PosKHG~\cite{PosKHG2023} further incorporates role-aware semantic modelling. This progression from aware-less to position-aware and role-aware modelling generally enables richer semantic representations and more expressive reasoning over $n$-ary relational facts~\cite{PosKHG2023,RAM2021}. A purely technique-oriented taxonomy would place these models in the same tensor-based category, despite their different semantic expressiveness and reasoning capabilities.

In contrast, our survey provides a broader analytical framework for $n$-ary relational representation learning.
We propose a two-dimensional taxonomy that jointly considers technical modelling principles and semantic awareness, distinguishing aware-less, position-aware, and role-aware models.
Specifically, it organises existing models based on (i) technical methodologies, i.e., tensor factorisation-based, translation-based, deep neural network-based, logic rule-based, and hyperedge expansion-based approaches, and (ii) levels of semantic awareness, i.e., aware-less, position-aware, and role-aware modelling.
Beyond taxonomy, we further discuss qualifier prediction, benchmark datasets, training settings, negative sampling strategies, benchmark-style model comparison, and open research challenges.
Table~\ref{tab:survey_comparison} summarises the main differences between our survey and existing related surveys.}

%%%%%%%%%%%%%%%%%%%%%%%%%%%%%%%%%%%%%%%%%%%%%%%%%%%%%%%%%%%%%%%%%%%%%%%%
\section{Two-dimensional Taxonomy for $n$-ary Relational Representation Learning Models}
\label{sec:taxonomy}
%%%%%%%%%%%%%%%%%%%%%%%%  TABLE  %%%%%%%%%%%%%%%%%%%%%%%%

\begin{table*}[!t]
\small
\centering
\caption{Two-dimensional Taxonomy for $n$-ary Relational Representation Learning Models.}
\label{tab:daxis}

% ---- local table spacing (only affects this table) ----
\setlength{\tabcolsep}{8pt}          % left/right padding inside cells (default ~6pt)
\renewcommand{\arraystretch}{1.15}   % row height multiplier
\setlength{\extrarowheight}{1.4pt}   % extra vertical padding

\begin{tabular}{|m{4cm}|m{3cm}|m{3cm}|m{3cm}|}
\hline
\textbf{} & \textbf{Role-aware} & \textbf{Position-aware} & \textbf{Aware-less} \\
\hline

\center\textbf{Translation-based} 
&   & m-TransH \cite{mtransH2016} & RAE \cite{RAE2018} \\
&   & BoxE \cite{boxe2020} &   \\
\hline

\center\textbf{Tensor factorisation-based}
& RAM \cite{RAM2021} & m-CP \cite{fatemi2019} & r-SimplE \cite{fatemi2019} \\
&  & HSimplE \cite{fatemi2019} & m-DistMult \cite{fatemi2019} \\
&  & HypE \cite{fatemi2019} & GETD \cite{GETD2020} \\
&  & ReAIE \cite{ReAIE2023} & S2S \cite{S2S2021} \\
&  & GAHE \cite{2025gahe} &  \\
&  & PosKHG \cite{PosKHG2023} &  \\
\hline

\center\textbf{Deep neural network-based} &  &   & \\
\hline

\multicolumn{1}{|c|}{{\small\itshape FCN-based}}
                    & NeuInfer \cite{neuinfer2020}  &   & \\
                    & NaLP \cite{NaLP2021} &   &\\
                    & t-NaLP \cite{NaLP2021} &  & \\
                    & ShrinkE \cite{ShrinkE2023}  &  & \\
\hline
\multicolumn{1}{|c|}{{\small\itshape CNN-based}}
                    &  HINGE \cite{HINGE2020} & HyConvE \cite{HyConvE2023}  & HyCubE \cite{2025hycube}\\  
                    &  & LPACN \cite{LPACN2023}  &\\
                    &  & HJE \cite{2024hje}  &\\
                    &  & HySAE \cite{2025hysae}  &\\

\hline
\multicolumn{1}{|c|}{{\small\itshape Transformer-based}}
                    & GRAN \cite{GRAN2021} & LKHGT \cite{lkhgt2025}  & \\
                    & HyNT \cite{HyNT2023} &  &\\
                    & Hy-Transformer \cite{HyTransformer2021}  &   & \\
                    & HyperMono \cite{2024hypermono}   &   & \\
                    & HyperFM \cite{2025hyperfm}   &   & \\
\hline
\multicolumn{1}{|c|}{{\small\itshape GNN-based}}
                    & STARE \cite{stare2020} & MAYPL \cite{maypl2025}  & \\
                    & STARQE \cite{StarQE2021} &   &\\
                    & QUAD \cite{QUAD2023} &   & \\
                    & HAHE \cite{HAHE2023} &   &\\
                    & HyperFormer \cite{hyperformer2023} &   &\\
                    & HypeTKG \cite{hyperTKG2023} &   &\\
                    & NE-Net \cite{NEnet2023} &   &\\
                    & HyperCL  \cite{2024hypercl}  &   &\\
\hline
\multicolumn{1}{|c|}{{\small\itshape HNN-based}}
                    & SDK \cite{SDK2023} & G-MPNN \cite{GMPNN2020}  & KHG-Aclair \cite{khg-aclair2024} \\
                    & RHKH \cite{2024rhkh} &H\textsuperscript{2}GNN \cite{H2GNN2024}  & Zhang et al. \cite{zhang2024}\\
                    &  & HYPER \cite{2025hyper}  & HyperQuery \cite{2025hyperquery}\\
\hline
% & NaLP \cite{NaLP2021} & HyConvE \cite{HyConvE2023} & ZHANG et al. \cite{zhang2024} \\
% & RHKH \cite{2024rhkh} & LPACN \cite{LPACN2023} & KHG-Aclair \cite{khg-aclair2024} \\
% & HINGE \cite{HINGE2020} & G-MPNN \cite{GMPNN2020} & HyCubE \cite{2025hycube} \\
% & NeuInfer \cite{neuinfer2020} & H\(^2\)GNN \cite{H2GNN2024} & HyperQuery \cite{2025hyperquery} \\
% & STARE \cite{stare2020} & HySAE \cite{2025hysae} &  \\
% & STARQE \cite{StarQE2021} &   & \\
% & GRAN \cite{GRAN2021} &   & \\
% & HyNT \cite{HyNT2023} &   & \\
% & QUAD \cite{QUAD2023} &   & \\
% & Hy-Transformer \cite{HyTransformer2021} &   & \\
% & ShrinkE \cite{ShrinkE2023} &   & \\
% & HyperMono \cite{2024hypermono} &   & \\
% & SDK \cite{SDK2023} &   & \\
% & HAHE \cite{HAHE2023} &   & \\
% & HyperFormer \cite{hyperformer2023} &   & \\
% & HyperTKG \cite{hyperTKG2023} &   & \\
% & NE-Net \cite{NEnet2023} &   & \\
% \hline

\center\textbf{Logic Rule-based} 
& HyperMLN \cite{HyperMLN2022} &   &  \\
\hline

\center\textbf{Hyperedge expansion-based} 
& TransEQ \cite{TransEQ2024} &   & \\
\hline
\end{tabular}
\end{table*}

To better understand and systematise existing approaches, we propose a two-dimensional taxonomy for \(n\)-ary relational representation learning methods. The first dimension categorises models by model techniques or principles, such as tensor factorisation-based methods, translation-based methods, deep neural network-based methods, and logic rules and hyperedge expansion-based methods. These approaches differ significantly in how they encode the semantic and relational information of \(n\)-ary facts.

\begin{itemize}
    \item \textbf{Translation-based methods.} Inspired by TransE-style models, these methods project entities and relations into a shared space and embed relations through entity projections and translations. For \(n\)-ary facts, scoring functions are extended over multiple entities.

    \item \textbf{Tensor factorisation-based methods.} Also known as tensor decomposition methods. These models represent \(n\)-ary relational facts as higher-order tensors and approximate them using low-rank decompositions, such as CP or Tucker. The goal is to capture higher-order interactions between entities while reducing the complexity of the \(n\)-ary representation.
    
    \item \textbf{Deep neural network-based methods.} These models leverage learnable architectures, e.g., Fully Connected Networks (FCNs), Convolutional Neural Networks (CNNs), Graph Neural Networks (GNNs), Hypergraph Neural Networks (HNNs), and Transformers to encode $n$-ary facts, often using shared or fact-level interaction functions. Their flexibility allows the modelling of arbitrary arity facts and complex relational patterns.
    
    \item \textbf{Logic rules and hyperedge expansion-based methods.} These models explore symbolic reasoning or structure transformation techniques to encode entity-role semantics, improving the interpretability of the model and modelling more complex hierarchical structures.
    
\end{itemize}

The second dimension reflects the degree to which models are aware of entity roles and entity positions in \(n\)-ary facts. On this basis we define three categories: position-aware models, role-aware models, and aware-less models.

% \begin{itemize}
    
%     \item \textbf{position-aware models} incorporate the relative or absolute position of each entity, typically via position embeddings or importance/weight of entities in \(n\)-ary facts.
    
%     \item \textbf{Role-aware models} go further by encoding entities and their specific roles in the relation, allowing fine-grained semantic representation of \(n\)-ary facts.

%     \item \textbf{Aware-less models} do not explicitly encode position or role information of entities within an \(n\)-ary facts. These models treat entities as an unordered tuple, reflecting only semantic information of entities and minimal higher-order structural information. 

% \end{itemize}

\begin{itemize}
    \item \textbf{Position-aware models} explicitly rely on the position index or relative position of entities within an $n$-ary fact, for example by using position embeddings, slot-specific projection matrices, or fixed weights for each position. Here, position serves only as a coarse proxy for semantic roles: the model can distinguish between ``the first'' and ``the third'' argument, but it does not introduce semantic role objects such as ``time'', ``location'', or ``participant'' that may be shared across different relations.
    \item \textbf{Role-aware models} explicitly encode semantic roles, such as relation-specific roles, qualifier relations, or role-value pairs. They introduce dedicated embeddings or parameters for roles, so that the same entity may contribute differently when it appears under different roles, and these roles can be shared across different relations (for example, the same ``time'' role can occur in multiple relations). If a model encodes both positional information and semantic roles, we also assign it to this category.
    \item \textbf{Aware-less models} do not explicitly encode either positional or role information for entities within $n$-ary facts. They treat the entities in a fact as an unordered (or only weakly ordered) collection, relying mainly on entity embeddings and minimal higher-order structural information.
\end{itemize}

% \textbf{\textit{Remark on the boundary between position-aware and role-aware models: }}
% In our terminology, role-aware models are those that introduce semantic roles as explicit, parameterised objects, whereas position-aware models rely on fixed argument slots as a coarse proxy. 
% For instance, although m-TransH \cite{mtransH2016} uses symbols $\rho^i$ to weight different arguments in an $n$-ary tuple, these $\rho^i$ are tied to the argument index within a specific relation and cannot be shared or reused as semantic roles across relations. We therefore regard m-TransH as position-aware rather than role-aware.

The boundary between position-aware and role-aware models may sometimes be subtle.
In this survey, we distinguish them according to whether the model introduces explicit semantic role representations.
Position-aware models distinguish entities by their positions within a relation, while role-aware models introduce semantic roles, qualifier relations, or role-value pairs as explicit modelling components.
For example, m-TransH~\cite{mtransH2016} assigns different weights to arguments through symbols such as $\rho^i$, but these symbols are tied to relation-specific 
positions rather than reusable semantic roles shared across relations.
We therefore classify m-TransH as position-aware rather than role-aware.
This two-dimensional taxonomy allows for a structured comparison of existing \(n\)-ary relational models. An overview of \(n\)-ary relational representation models under this taxonomy is presented in Table~\ref{tab:daxis}.

% #######################################################################
% #######################################################################
\subsection{Position-aware Models}
\label{sec:ordered}

% position-aware models incorporate information about the position or importance of entities within an \(n\)-ary fact. By using position encoding, dedicated positional transformation functions or entity weights, these models can implicitly learn the role contributions through position-aware embeddings.

Position-aware models distinguish entities according to their positions or slots within an $n$-ary fact, for example by using position embeddings, slot-specific transformations, or position-dependent weights.

% #######################################################################
\subsubsection{Translation-based Methods} 

We focus on \emph{translation-based} approaches within the \emph{position-aware} category. 
Here, positional information is typically introduced by associating the positional indices or position-dependent weights of entities, while retaining a translation-style scoring function.

\textbf{m-TransH} \cite{mtransH2016} is an early translation-based model designed for \(n\)-ary relational representation learning. 
It extends TransH by modelling an \(n\)-ary fact as a relation-specific assignment of entities to argument slots, and associates each relation \(r\) with two orthogonal unit vectors \(n_r\) and \(b_r\), together with a weighting function \(a_r(\rho)\) that modulates the contribution of each argument position \(\rho\). 
The scoring function is defined in Eq.~\ref{eq:mtransh}: 
\begin{equation} f_r(t) = \Big|\sum_{\rho\in \mathcal{M}(R_r)} a_r(\rho) , P_{n_r}(t(\rho)) + b_r \Big|^2, \label{eq:mtransh} \end{equation}
where \(P_{n_r}(z)\) denotes the projection of the vector \(z\) onto the hyperplane \(n_r\). 
Although m-TransH uses the term ``role'', these roles correspond to schema-defined entity/argument positions that are specific to each relation, rather than to explicit semantic roles that can be shared across relations. 
From a modelling perspective, m-TransH is not fully expressive (see Def.~\ref{def:fullexp}), which limits its capacity to represent complex relational patterns \cite{SimplE2018,stare2020,ReAIE2023}.

To address the expressiveness limitation of m-TransH, \textbf{BoxE} \cite{boxe2020} proposes a fully expressive translation-based formulation for \(n\)-ary facts. 
BoxE represents entities as points and relations as hyper-rectangles (boxes) and assumes an \(n\)-ary fact is valid if the translated embedding of each entity lies within its corresponding relation-specific box. 
Although BoxE can model a broader class of relational patterns, such as symmetry and inversion, it still relies on entity positions rather than explicit, shareable semantic roles, thus remaining within the position-aware category \cite{HyperMLN2022}.

% To overcome this limitation, \textbf{BoxE} \cite{boxe2020} is proposed as a fully expressive translation-based model. It represents entities as points and relations as hyper-rectangles (or boxes). Each entity is parameterised by a base vector \(\mathbf{e}_i\) and a translation vector \(\mathbf{b}_i\), while an \(n\)-ary fact is valid if each translated entity embedding falls within its corresponding box. The model supports logical properties such as symmetry and inversion, but has limitations in capturing complex patterns \cite{HyperMLN2022}.

% #######################################################################
\subsubsection{Tensor factorisation-based Models}

In the \emph{position-aware} setting, tensor factorisation-based models typically encode positional distinctions by assigning position-dependent embeddings or transformations to entities, and aggregate them through a multilinear scoring function, thus enabling entities to contribute differently across argument slots.

\textbf{m-CP} \cite{fatemi2019} is one of the first attempts to extend tensor factorisation-based KGRL from binary relations to the general $n$-ary setting in a position-aware manner.
It generalises the classical CP decomposition \cite{CP1927} by assigning \(\alpha\) different embeddings to each entity, where \(\alpha\) is the maximum arity in the dataset. 
For a fact \(r(e_1,\dots,e_\alpha)\), the scoring function aggregates the position-specific embeddings via a multilinear product:
\begin{equation}
\phi\big(r(e_1,\dots,e_\alpha)\big) 
  = \odot(\mathbf{r}, \mathbf{e_1}, \dots, \mathbf{e_\alpha}).
\label{eq:mcp}
\end{equation}
This construction enables m-CP to distinguish entities by their argument positions, but it also means that the same entity is represented by unrelated vectors in different slots, leading to parameter redundancy and poor generalisation to mixed-arity datasets \cite{GRAN2021}.

To alleviate the lack of shared semantics across positions in m-CP,  
\textbf{HSimplE} \cite{fatemi2019} retains position-awareness while enforcing a more compact parameterisation than m-CP. It assigns a single base embedding to each entity and derives position-specific representations via a deterministic shift operator. 
This encourages information sharing across positions, but the hard-coded shift limits the richness of position-dependent semantics and motivates the use of learnable positional convolutions in \textbf{HypE} \cite{fatemi2019}. 

HypE replaces the fixed shift operator with a learnable convolutional transformation \(\mathbf{f}(\mathbf{e}, i)\) that jointly considers the entity embedding and its positional index. 
For a tuple \((r,e_1,\dots,e_n)\), the scoring function of HypE is:
\begin{equation}
\phi\big(r(e_1,\dots,e_n)\big) 
  = \odot\big(\mathbf{r}, \mathbf{f}(\mathbf{e_1}, 1), \dots, \mathbf{f}(\mathbf{e_n}, n)\big).
\label{eq:hype}
\end{equation}
By learning position-specific filters, HypE achieves full expressiveness and provides a more flexible mechanism for modelling positional variation compared with HSimplE. 
However, its convolutional mechanism increases the model complexity compared with HSimplE.

While m-CP, HSimplE, HypE are formulated in a KHG setting, recent work \textbf{GAHE} \cite{2025gahe} continues the line of position-aware tensor factorisation models by integrating multi-curvature geometric representations in the HKG setting, where primary triples together with qualifiers give rise to richer geometric structures, such as chains, rings, and hierarchical patterns. 
GAHE combines a weighted multilinear scoring function with geometry-aware embeddings defined over Euclidean, hyperbolic and spherical spaces, and introduces position-dependent sub-relations so that different qualifier slots can be modelled in dedicated subspaces. 
This geometry-augmented design enables GAHE to better capture mixed structural patterns in HKG, while retaining full expressiveness and reducing the dimensionality required compared with earlier HKG models such as RAM \cite{RAM2021} and S2S \cite{S2S2021}. However, its multi-curvature and subspace operations require more implementation effort compared with purely Euclidean tensor factorisation models.

Orthogonal to the above line of work, \textbf{ReAIE} \cite{ReAIE2023} focuses on relational-algebra-style reasoning rather than on new parameterisations or geometries. 
It is a fully expressive tensor factorisation KHG model that partitions embeddings into local windows and applies weighted aggregation, so that it can implement a broad subset of primitive relational algebra operations such as renaming, projection, set union, selection, and set difference. 
In our taxonomy, ReAIE still falls within the position-aware category, because it relies on position-specific contributions and does not introduce explicit semantic roles.

% #######################################################################
% \subsubsection{Random Walk Based Methods}

% #######################################################################
\subsubsection{Deep Neural Network-based Methods}

Deep neural network-based position-aware models encode the \emph{index of an entity in a relation} via learnable architectures, such as CNNs, Transformers and HNNs, so that entity representations and interactions become explicitly position-dependent.

\paragraph{CNN-based models}

% To move beyond the local triple–qualifier quintuples used in HINGE, 
\textbf{HyConvE} \cite{HyConvE2023} is introduced as the first CNN-based model that models each $n$-ary fact as a whole tuple with CNNs.
The model reshapes entity and relation embeddings from \(\mathbb{R}^d\) to 2D matrices in \(\mathbb{R}^{d_1 \times d_2}\) with \(d_1 d_2 = d\), stacks these matrices into a 3D tensor, and applies 3D convolutions to capture interactions among all arguments.
It further uses position-specific convolutional filters so that entities at different indices follow distinct convolutional paths, which makes the model position-aware in our taxonomy.
This design improves the modelling of higher-order interactions, but the position-specific filters introduce many extra parameters and limit scalability to high-arity and mixed-arity graphs.

To model the contribution of each entity in an $n$-ary tuple more explicitly than HyConvE, \textbf{LPACN} \cite{LPACN2023} adopts a CNN architecture that is inspired by the position-aware design of HypE. It adds an attention mechanism on top of convolutional features to aggregate entity embeddings into a relation-aware representation and encodes entity indices through convolutional kernels.
However, the combined CNN–attention stack often leads to gradient vanishing; the extended variant \textbf{LPACN+} adds residual connections to stabilise model training, at the price of a more complex architecture.

To address two specific limitations of HyConvE, namely the lack of positional encoding in the 3D convolutional path and the reliance on separate position-specific filters in the 2D paths, \textbf{HJE} \cite{2024hje} proposes a joint convolutional architecture for KHG completion. It introduces a unified learnable position matrix and adds position embeddings to entities at each index before both 3D and 2D convolutions. Interaction enhanced 3D convolutions and relation-aware 2D convolutions then jointly extract local and global patterns from the stacked relation–entity tensor. 
Like other convolutional models, it still retains a comparatively complex architecture.

Since CNN-based KHG models such as HyConvE, LPACN and HJE still rely on heavy convolutional structures with many parameters, \textbf{HySAE} \cite{2025hysae} aims to improve efficiency and scalability. It proposes a semantic-enhanced 3D scalable architecture that represents each fact as a 3D tensor and applies lightweight 3D convolutions to capture higher-order interactions. Position embeddings are explicitly merged with entity embeddings to form a position-enhanced 3D tensor, which is then passed through internal and external semantic enhancement layers to perform multi-scale semantic aggregation. 
As a result, HySAE achieves competitive performance with a smaller parameter budget, although it still inherits the intrinsic limits of CNNs, such as a receptive field constrained by kernel size and depth.

\paragraph{Transformer-based models}

\rev{Most Transformer-based models for $n$-ary relational data focus on modelling interactions within hyper-relational facts or improving link prediction. However, complex query answering over KHGs remains less explored, especially when queries involve logical operators over ordered hyperedges. }
To address this gap, \textbf{LKHGT} \cite{lkhgt2025} proposes a two-stage Transformer architecture, with a particular focus on \textit{EFO-1 queries}, i.e., existential first-order queries with a single free variable built from projection, conjunction, disjunction, and negation.
A Projection Encoder first performs atomic projections over ordered hyperedges, while a Logical Encoder composes these projected embeddings using logical operators such as conjunction, disjunction, and negation.
% A Projection Encoder first performs atomic projections over ordered hyperedges, while a Logical Encoder composes these projected embeddings using first-order queries with a single free variable logical operators such as conjunction, disjunction, and negation. 
This design enables LKHGT to answer EFO-1 queries over KHGs and to generalise to out-of-distribution query structures, but it still assumes static settings and focuses on single-variable queries.

\paragraph{GNN- and HNN-based models}

Many existing HKG models either treat each fact independently or rely on shallow neighbourhood structures, making them unsuitable for inductive reasoning. Furthermore, they often underutilise the structural information inherent in HKGs, especially the position of entities and relations within each fact. 
To address these limitations, \textbf{MAYPL} \cite{maypl2025} is proposed as a unified structure-driven GNN framework that fully leverages position-aware patterns for both transductive and inductive link prediction. 
It learns initial entity and relation embeddings via a structure-driven initializer, which exchanges messages between co-occurring entities and relations using position-specific projections (e.g., head, tail, qualifier). These representations are further refined through attentive message passing and updated via both entity-centric and relation-centric aggregation schemes. 
Its main drawback is the computational cost introduced by complex attention mechanisms, and the model remains limited to static settings.

While most early HNN models overlooked positional semantics, \textbf{G-MPNN} \cite{GMPNN2020} presents one of the first attempts to encode entity positions within hyperedges. It models a KHG as a multi-relational ordered hypergraph and performs generalised message passing over labelled and indexed hyperedges. The index of each entity is encoded explicitly through a positional mapping $P_e$, which is passed to the message and aggregation functions, making the model position-aware. Nevertheless, G-MPNN assumes symmetric relations and lacks the capacity to capture deeper structural semantics \cite{ReAIE2023}.

To incorporate hierarchical semantics and geometric inductive bias, \textbf{H}$^2$\textbf{GNN} \cite{H2GNN2024} reformulates KHGs into tree-like structures by instantiating $n$ auxiliary hyperedges for each $n$-ary fact. The model then performs two-stage message passing in hyperbolic space using Lorentz transformations, which naturally reflect hierarchical distances. Positional information is preserved by the decomposition of hyperedges and the use of hierarchy-aware aggregation in the Lorentzian manifold. 
Compared to G-MPNN, this design allows the model to reflect both local roles and global structure more effectively.

To overcome the geometric rigidity and limited reasoning depth of H$^2$GNN, \textbf{HYPER} \cite{2025hyper} proposes a hybrid architecture that disentangles local and global reasoning over KHGs. In the local stage, it applies position-aware convolutional encoders over entity-position pairs using shared kernels. In the global stage, hyperedge-level attention with positional bias enables the model to perform multi-hop reasoning and infer long-range dependencies. Unlike H$^2$GNN, which relies on geometric hierarchies, HYPER introduces position-sensitive feature disentanglement and flexible attention-based aggregation, achieving stronger reasoning capacity over positional semantics.
Although HYPER achieves expressive modelling of position-aware semantics through its architecture, the combination of local convolutions and global attention introduces additional computational cost, potentially limiting its scalability to very large or high-arity hypergraphs.

% #######################################################################
% #######################################################################
\subsection{Role-aware Models}
\label{sec:roles}
Role-aware models explicitly encode the semantic roles that entities play in a relation. Rather than relying solely on entity position, these models assign role-specific embeddings or projections, allowing them to distinguish between entities of the same type, or the same entity in different positions with different contextual meanings. This fine-grained modelling improves relational expressiveness, especially in \(n\)-ary facts.
Note that if the model encodes both positional and role information, it also falls into this category.

% #######################################################################
% \subsubsection{Translation Based Methods}

% #######################################################################
\subsubsection{Tensor factorisation-based Models}

% To model \(n\)-ary relational semantics more precisely, role-aware models based on tensor factorisation methods explicitly encode role-specific embeddings for each entity and capture fine-grained entity-role compatibility.

In the role-aware setting, tensor factorisation-based models introduce dedicated embeddings for semantic roles and use multilinear scoring functions to model the compatibility between entities and roles. 
Compared with position-aware tensor models, they find role embeddings rather than relying solely on entity positions.

\textbf{RAM} \cite{RAM2021} is one of the first tensor factorisation-based models that treats roles as first-class objects in KHGs. 
It is introduced to improve the expressiveness and efficiency of \(n\)-ary relational modelling by learning semantic embeddings for roles, instead of treating all arguments as anonymous positions or flat attributes. 
RAM constructs a latent role space and represents each role of a relation as a linear combination of these basis vectors, so that similar roles share similar semantic representations. 
Each role is further associated with a learnable pattern matrix \(\mathbf{P}_i^r\), which models the compatibility between relation \(r\) and the entities playing its \(i\)-th role. 
The plausibility of a fact is then computed by a multilinear scoring function:
\begin{equation}
\phi(x) = \sum_{i=1}^{a_r} \left\langle \mathbf{u}_i^r, \mathbf{P}_i^r[1,:] \mathbf{E}_1, \cdots, \mathbf{P}_i^r[a_r,:] \mathbf{E}_{a_r} \right\rangle,
\label{eq:ram}
\end{equation}
where \(\mathbf{u}_i^r\) is the embedding for the \(i\)-th role of relation \(r\), \(a_r\) is the arity of \(r\), and \(\mathbf{E}_i\) maps each entity to multiple semantic embeddings.
RAM is fully expressive and explicitly role-aware, but it does not encode positional information, and it treats entity–role pairs in a relatively uniform way without explicitly distinguishing their relative contributions \cite{ShrinkE2023}.

\textbf{PosKHG} \cite{PosKHG2023} builds directly on RAM and is introduced to further enhance its expressiveness by incorporating positional information in KHGs.
It enhances RAM’s role-aware representations with a hard-coded position shift, in the spirit of HSimplE: entity semantic embeddings are shifted in different argument slots to obtain distinct position-dependent embeddings.
This design strengthens the model’s ability to distinguish argument configurations while adding very little extra computational complexity compared with RAM.

% \rev{Viewed together, RAM and PosKHG trace a concise progression within role-aware tensor factorisation for KHGs: RAM firstly treats roles as explicit parameters in a multilinear scoring framework, and PosKHG further enriches this formulation by adding lightweight, HSimplE-style position shifts. At the same time, both models remain confined to static KHG link prediction and rely only on symbolic relational structure and schema-level role annotations; they do not address inductive or temporal settings, nor do they incorporate textual or multimodal features or deeper neural architectures for role modelling.}

% #######################################################################
\subsubsection{Deep Neural Network-based Models}

% Role-aware models based on deep neural networks capture the role semantics of entities in \(n\)-ary facts, typically by modelling entity-role pairs. There are three main cases: (i) facts are represented as tuples of role-entity pairs; (ii) facts are represented as HKGs, where qualifiers are sets of role-entity pairs; and (iii) models that encode each role semantic in the \(n\)-ary fact from the representation of the primary relation.

Role-aware deep neural network-based models explicitly encode the semantic roles that entities play in $n$-ary facts, typically by treating entity–role pairs as basic modelling units. Unlike position-aware models, which distinguish entities mainly by their indices within a fact, these methods learn dedicated role or qualifier embeddings and model interactions between roles and entities through various neural architectures, including FCNs, CNNs, Transformers, GNNs and HNNs.
% In this way, the same entity can behave differently under different roles, and the same role can be shared across multiple relations, enabling richer and more fine-grained representations of \(n\)-ary semantics.

\paragraph{FCN-based models}

\textbf{NeuInfer} \cite{neuinfer2020} is introduced to better exploit the distinction between a primary triple and its auxiliary descriptions in hyper-relational facts. It can support both simple inference on complete facts and flexible inference on partially observed facts.
% Each \(n\)-ary fact is represented as a primary triple \((h,r,t)\) together with a set of attribute–value pairs \(\{(a_i,v_i)\}\). 
NeuInfer uses FCNs to compute a validity score for the primary triple and a compatibility score between the triple and each qualifier pair. 
Roles are explicitly represented by embeddings of relations and attributes in qualifiers, so that the same entity can contribute differently under different attributes. 
Subsequent work has also pointed out that NeuInfer learns roles largely in isolation, offers only limited modelling of dependencies among auxiliary descriptions within an $n$-ary fact, and requires many parameters that make it prone to over-fitting \cite{RAM2021,GRAN2021,ShrinkE2023}.

In contrast to NeuInfer, which singles out a primary triple, \textbf{NaLP} \cite{NaLP2021} treats each $n$-ary fact as an unordered set of role–value pairs $rel\{r_1\!:\!v_1,\dots,r_m\!:\!v_m\}$ and aims to model the relatedness among these pairs directly. 
It encodes each \((r_i,v_i)\) pair with a FCN and then uses another FCN to estimate pairwise relatedness between all pairs, followed by permutation-invariant aggregation to obtain a fact-level representation and score. 
Roles are then explicitly encoded through dedicated role embeddings at the relation-value level. 
The extended \textbf{tNaLP} model further introduces a type-compatibility branch that constrains each role–value pair with learned type embeddings, while NaLP+ and tNaLP+ improve the negative sampling strategy for harder negatives.
Thus, NaLP-based models can support mixed-arity facts, but they are not fully expressive and cannot differentiate major and minor role-entity pairs \cite{stare2020,ShrinkE2023}.

While NeuInfer and NaLP-based models mainly learn compatibility among triples, qualifiers, or role-value pairs, \textbf{ShrinkE} \cite{ShrinkE2023} further focuses on logical inference patterns in hyper-relational facts. 
It represents the primary triple as a relation-specific query box and models each qualifier as a box-shrinking operation that narrows the possible answer set. 
The shrinking function is learned with a FCN from the primary relation, qualifier relation, and qualifier value, enabling ShrinkE to capture qualifier monotonicity, implication, and mutual exclusion. 
Although ShrinkE does not introduce explicit role embeddings in the same way as role-value-pair models, it is role-aware in our taxonomy because it explicitly models qualifier relations and values. 
Its main limitation is the monotonicity assumption, namely that adding qualifiers should shrink the answer set, which may not hold for semantically opaque or context-dependent qualifiers.

% While \textbf{BoxE} encodes multi-answer queries via geometric translations and rotations of entities and relations in a box space, it suffers from high computational overhead due to complex geometric rules and a large number of parameters. Moreover, it struggles to naturally model the semantics of ambiguous or logically complex qualifiers. To address these issues, \textbf{ShrinkE} \cite{ShrinkE2023} simplifies the geometric structure and focuses on logical expressivity. It abandons translation-based operations and uses FCNs to directly learn "shrink" functions, which encode logical properties of qualifiers such as monotonicity, implication, and mutual exclusion. This design reduces model complexity while allowing more flexible constraint modelling. However, ShrinkE does not encode role or positional semantics, and its monotonicity assumption, that adding qualifiers will always reduce the answer set, limits its applicability to qualifiers with opaque or context-dependent meanings \cite{ShrinkE2023}.

\paragraph{CNN-based models}
\textbf{HINGE} \cite{HINGE2020} is an early CNN-based HKG model that aims to exploit qualifiers while still preserving the central role of the primary triple in hyper-relational facts. It assumes a strong correlation between a valid primary triple and its associated qualifiers, and therefore learns interaction features both from the primary triple $(h,r,t)$ and from quintuples $(h,r,t,k_i,v_i)$. Each triple or quintuple is reshaped into a 2D “image” where the rows correspond to fixed positions such as head, relation, tail, key and value. This fixed ordering, together with position-sensitive CNN filters, makes HINGE role-aware in our taxonomy. 
However, HINGE models each fact locally, captures role–entity interactions independently, and relies only on static structural information without deeper message passing across facts \cite{stare2020,RAM2021,GRAN2021}.

\paragraph{GNN-based models}

\textbf{STARE} \cite{stare2020} is one of the first GNN- and Transformer-based models designed for link prediction on HKGs. 
It adopts an encoder-decoder architecture, using CompGCN \cite{CompGCN2019} to update entity and relation embeddings through message passing over hyper-relational facts, and a Transformer-based decoder for missing entity prediction. 
Roles are encoded by distinguishing the primary relation, qualifier relations, and qualifier values, so that the same entity may contribute differently when appearing in different parts of a hyper-relational fact. 
However, STARE models qualifier information in a relatively coarse way, without explicitly distinguishing the importance of different qualifiers or modelling fine-grained interactions between qualifiers and the primary triple \cite{GRAN2021,QUAD2023}.

\rev{Building on STARE, \textbf{QUAD} \cite{QUAD2023} further improves the interaction modelling between primary triples and qualifiers. 
While previous HKG encoders mainly emphasise how qualifiers enhance the representation of base triples, QUAD additionally models the reverse information flow from base triples to qualifiers. 
It introduces a dual-aggregator architecture: one aggregator updates base-triple representations using their associated qualifiers, while the other updates qualifier representations using the corresponding base triples. 
A Transformer-based decoder then models the interactions between these two views for missing entity and qualifier prediction. 
% QUAD is role-aware because it explicitly distinguishes primary relations, qualifier relations, and qualifier values, and learns their bidirectional dependencies within each hyper-relational fact. 
This design improves qualifier representations and strengthens HKG completion, but the dual aggregators and Transformer decoder introduce additional parameters and may increase the risk of over-fitting \cite{ShrinkE2023}.
}

% StarE \cite{stare2020} encodes hyper-relational facts using triple-centric aggregation but fails to fully capture the influence of base triples on qualifiers. To address this, \textbf{QUAD} \cite{QUAD2023} builds on StarE and introduces a dual-aggregator architecture.
% One encoder aggregates qualifiers around each base triple, while the other aggregates base triples around each qualifier. Then a Transformer-based decoder models interactions between these two views to predict missing entities and qualifiers. This bidirectional design improves the quality of qualifier representations and leads to stronger HKG link prediction. However, the dual encoders and Transformer decoder form a complex architecture with many parameters, which increases the risk of over-fitting \cite{ShrinkE2023}.

While STARE mainly targets link prediction, \textbf{STARQE} \cite{StarQE2021} extends the STARE encoder towards complex query answering. 
It reuses a STARE-style hyper-relational encoder to obtain role-aware embeddings for entities, relations, and qualifiers, and introduces a graph-attention-based query encoder with logical symbols for multi-hop reasoning over HKGs. 
STARQE is therefore an early attempt to extend GNN-based role-aware HKG models from link prediction to CQA, although it does not support literals in qualifiers and only targets a restricted fragment of first-order queries.

Several later models further refine how qualifier information interacts with the primary triple. 
\textbf{HyperFormer} \cite{hyperformer2023} addresses the noise introduced by global multi-hop aggregation in models such as STARE and QUAD by focusing on local-level sequential information. 
It introduces an entity neighbour aggregator, a relation qualifier aggregator, and a convolution-based bidirectional interaction module to capture interactions among entities, relations, and qualifiers. 
In particular, the relation qualifier aggregator incorporates qualifier relations and qualifier entities into relation representations, making HyperFormer role-aware under our taxonomy. 
Although this local interaction design improves hyper-relational modelling, the model remains computationally demanding due to its Transformer and interaction modules.

\textbf{HAHE} \cite{HAHE2023} combines global and local modelling through a hierarchical attention architecture. 
At the global level, it views the HKG as a hypergraph and applies hypergraph dual-attention layers to aggregate information between entities and hyperedges. 
At the local level, it linearises each hyper-relational fact into a semantic sequence consisting of the primary triple and qualifiers to capture role-specific interactions. 
This design yields role-aware representations at both graph and fact levels and supports multi-position prediction. 
Nonetheless, HAHE is most effective on datasets with rich hyper-relational facts, and later multimodal work shows that structure-only HKG embeddings can be suboptimal when textual and visual features are available \cite{2025hyperfm}.

\textbf{HyperCL} \cite{2024hypercl} further improves GNN-based HKG models from an ontology-aware contrastive learning perspective. 
It observes that existing HKG encoders, such as STARE, Hy-Transformer, QUAD, and HAHE, mainly rely on instance-level hyper-relational facts and often overlook the hierarchical ontology of entities. 
HyperCL keeps the original HKG encoder--decoder pipeline as an instance view, and builds an ontology view where entities and concepts are connected by \emph{instance-of} and \emph{subclass-of} relations and encoded with relation-aware GATs. 
A concept-aware contrastive loss then aligns the representations of the same entity across the two views while separating semantically close but distinct entities. 
As a plug-in framework, HyperCL inherits the role-aware modelling ability of its base HKG encoders and consistently improves entity prediction, but it relies on reasonably complete ontology information and increases training complexity due to the additional GAT and contrastive learning objective.

% Finally, GNN-based role-aware modelling has also been extended to temporal settings. 
Finally, building on GNN-based role-aware HKG modelling such as STARE~\cite{stare2020} and its follow-up models, recent work has also extended this line of research to temporal settings.
\textbf{HypeTKG}~\cite{hyperTKG2023} introduces HTKGs by adding timestamps to HKG facts and learns time-aware role-sensitive representations for temporal link prediction, whereas \textbf{NE-Net}~\cite{NEnet2023} proposes an $n$-ary temporal KG structure to model evolving entity-role pairs across time snapshots.
Both models are discussed in Section~\ref{sec:temporal_ext}.

% \textbf{HypeTKG} introduces HTKGs by adding timestamps to HKG facts and learns time-aware role-sensitive representations for temporal link prediction, whereas \textbf{NE-Net} proposes an $n$-ary temporal KG structure to model evolving entity--role pairs across time snapshots. 

\paragraph{Transformer-based models}

To overcome the coarse treatment of qualifiers in STARE and to better capture dependencies within a hyper-relational fact, \textbf{GRAN} \cite{GRAN2021} models each fact as a small heterogeneous graph. 
This local graph view provides an early example of representing a hyper-relational fact as a structured fact-level object, which is related to later hypergraph-style formulations that jointly encode the primary triple and qualifiers.
It introduces a Transformer-based multi-head attention mechanism with edge-aware attention biases that captures detailed interactions between relations and entities, including those in qualifiers. In this way, the same entity is embedded differently when it appears as subject, object or qualifier value. 
Nonetheless, GRAN uses a fully connected attention mechanism over all nodes in each fact, which leads to a large number of parameters, and it operates only on discrete entities \cite{ShrinkE2023,HyNT2023}.

\textbf{HyNT} \cite{HyNT2023} further targets HKGs with numerical literals. It encodes triples and qualifiers with shared neural modules, and then uses a context Transformer with distinct position embeddings to distinguish triple tokens from qualifier tokens. Numerical literals such as time and quantity are mapped into the embedding space, and a prediction Transformer is used to handle discrete entity prediction, relation prediction and numerical regression within a unified framework. HyNT is therefore role-aware through its use of relation- and qualifier-specific parameters. 
Nevertheless, it still relies on stacked Transformer blocks and per-relation numeric projection parameters, which leads to a considerable computational overhead, especially in HKGs with numerous relations and qualifiers, and its linear numeric projection limits the ability to capture more complex non-linear numerical patterns.

\rev{Beyond numerical literals, \textbf{HyperFM} \cite{2025hyperfm} extends Transformer-based HKG link prediction to the multimodal setting. 
It introduces a fact-centric hypergraph Transformer that jointly encodes structural, visual, and textual features associated with hyper-relational facts. 
Edge-biased self-attention is used to distinguish relation and modality types, allowing the model to learn modality-aware and role-sensitive representations. 
HyperFM is therefore role-aware in our taxonomy, since qualifier-related relations and modality-specific signals are explicitly used to refine the representation of each hyper-relational fact. 
However, it still relies on a relatively heavy Transformer architecture and does not explicitly address missing modalities.}

Unlike the above Transformer-based HKG models, which prioritise representation power, \textbf{Hy-Transformer} \cite{HyTransformer2021} mainly targets computational efficiency and scalability. It removes GNN-based encoders and replaces them with lightweight embedding layers and keeps a Transformer-based decoder over hyper-relational statements. In addition, a qualifier-masking auxiliary task is introduced to strengthen supervision on qualifiers and improve qualifier prediction. 
This simplified architecture reduces computational cost, but the lighter encoder risks losing more subtle structural information compared with models that explicitly aggregate graph neighbourhoods \cite{GRAN2021}.

\rev{Existing Transformer-based HKG models mainly improve interaction modelling between primary triples and qualifiers, but they usually do not explicitly enforce the monotonic effect of qualifiers. 
To address this limitation, \textbf{HyperMono} \cite{2024hypermono} introduces a monotonicity-aware Transformer architecture based on two principles: \emph{stage reasoning} and \emph{qualifier monotonicity}. 
It first derives coarse predictions from the primary triple, and then refines them with a qualifier-monotonicity-aware cone module, where the primary triple defines a cone and each qualifier acts as a shrinker that narrows the possible answer set. 
HyperMono is role-aware in our taxonomy because qualifier relations and values are explicitly used to refine the semantics of the primary triple. 
This design combines Transformer-based expressiveness with monotonicity-guided reasoning, but it may still incur notable computational cost when qualifier information is sparse or weak.
}

\paragraph{HNN-based models}

Motivated by the need to handle noisy, sparse and dynamically evolving interactions in recommender systems built on HKGs, \textbf{SDK} \cite{SDK2023} extends the STARE-style hyper-relational modelling to a dynamic HKG setting with self-supervision.
Each interaction is represented as a primary triple augmented with contextual qualifiers, and dynamic hypergraph convolution together with self-attention are applied to propagate information over time. 
Different node and edge types (users, items, contextual entities and relations) are associated with distinct parameters, which makes the learned embeddings role-aware while benefiting from the hypergraph structure. 
However, its dynamic setting and self-supervised objectives lead to a relatively complex architecture that is tailored to recommendation rather than general HKG completion \cite{SDK2023}.

\textbf{RHKH} \cite{2024rhkh} takes a different perspective and targets general LP directly on KHGs. Instead of decomposing \(n\)-ary facts into triples or role–value sets, it preserves each fact in its native form \(r(e_1,\dots,e_{|r|})\) and models the resulting structure as a relational hypergraph. For each hyperedge, the relation \(r\) is decomposed into role-specific vectors \(\{r_1,\dots,r_{|r|}\}\), which interact with entities at corresponding positions. 
Position-dependent shift functions \(f^p_{r_i}(\cdot)\) further distinguish positional semantics.
% While position-dependent shift functions \(f^p_{r_i}(\cdot)\) further distinguish positional semantics. 
RHKH therefore jointly encodes role, position and structural information, enabling fine-grained modelling of intra- and inter-tuple dependencies in KHGs. 
However, its hierarchical aggregation and convolution-based scoring still incur relatively high computational cost.

% #######################################################################
\subsubsection{Logic Rules and Hyperedge expansion-based Models}

Several papers explore alternative techniques to encode role semantics via logic rules or hyperedge expansion \cite{HyperMLN2022,TransEQ2024,HGsurvey2023}.
\textbf{HyperMLN} \cite{HyperMLN2022} is a hybrid framework that combines KHG embeddings with Markov Logic Networks (MLNs) \cite{MLN2006}. It introduces an \(n\)-ary facts-adapted MLN that avoids the star-to-clique transformation, and captures role semantics via weighted first-order logic rules. A variational EM algorithm jointly optimises the KHG embeddings and logic rule weights: the E-step effectively distils the knowledge into KHG embeddings; the M-step updates the rule weights and the model parameters. HyperMLN achieves both predictive performance and interpretability, but is sensitive to the coverage and quality of the mined logic rules \cite{HyperMLN2022}.
% approximates the posterior distribution over hidden tuples by extracting their Markov blanket, maximizes the joint likelihood to 

\textbf{TransEQ} \cite{TransEQ2024} is the first fully expressive model using the equivalent transformation. It transforms HKGs into KGs by introducing a mediator entity and sub-relations, while fully preserving the original semantics and structure of HKGs. The model adopts an encoder-decoder framework, where the GNN-based encoder learns structural representations and the decoder captures semantic information. Its scoring function is given in Eq.~\ref{eq:transeq}: 
\begin{equation} 
\phi(x)=\langle \psi(h_r^L, h_{a_1}^L, \dots,h_{a_n}^L), h_{s}^L, h_{o}^L, h_{v_1}^L, \dots,h_{v_n}^L \rangle, 
\label{eq:transeq}
\end{equation} 
where \(L\) is the number of GNN layers, \(\psi\) denotes the aggregation function, and \(\langle \cdot \rangle\) is the multilinear product.

HyperMLN highlights the potential of logic rule-based methods to improve model interpretability. TransEQ, on the other hand, demonstrates the effectiveness of the hyperedge expansion method in learning complex higher-order and hierarchical structures.

% % #######################################################################
% % \subsubsection{Reification Based Models}

% \textbf{TransEQ} \cite{TransEQ2024} is the first fully expressive model to use an equivalent transformation technique. It transforms hyper-relational knowledge graphs (HKGs) into KGs while maintaining semantic and structural integrity by adding a mediator entity \(b_k\) and two sub-relations \(r^{sub}\) and \(r^{obj}\). The model then makes use of an encoder-decoder framework, where the encoder uses GNNs to capture the structural information of the KG and the decoder models semantic information using HKG scoring functions (e.g., SF of m-DistMult \cite{fatemi2019}). The scoring function is reconstructed as \(\phi(x)=\langle \psi(h_r^L, h_{a_1}^L, \dots,h_{a_n}^L), h_{s}^L, h_{o}^L, h_{v_1}^L, \dots,h_{v_n}^L \rangle\), with \(L\) standing for the number of GNN layers in the encoder, \(\psi\) for the aggregation function, and \(\langle\cdot\rangle\) for the multilinear product.

% #######################################################################
% #######################################################################
\subsection{Aware-less Models}
\label{sec:unordered}

Aware-less models treat the entities in an \(n\)-ary relation without explicitly modelling their positions or semantic roles. 
They do not distinguish how much each entity contributes to the fact based on its position or role. 
Any ordering that appears in the data is typically incidental to how the dataset is constructed, rather than the result of an explicit positional encoding in the representation learning process. 
This category offers simplicity and efficiency, but may lack the ability to capture fine-grained relational semantics.

% #######################################################################
\subsubsection{Translation-based Methods}

Translation-based aware-less models represent entities and relations by geometric translations in the embedding space, without attaching explicit positional or role parameters to the arguments of an $n$-ary fact.

\textbf{RAE} \cite{RAE2018} extends the translation-based paradigm to \(n\)-ary relations with the specific goal of making knowledge base completion scalable in the instance reconstruction setting. It models each \(n\)-ary fact as a fixed-order tuple and adopts the m-TransH embedding loss, while introducing an additional relatedness loss that estimates how likely two entities are to co-occur in a common instance. Although entity order is preserved in the data, RAE does not assign separate embeddings to different positions or semantic roles, so in our taxonomy it remains an aware-less model. 
Subsequent work shows that this design improves LP and instance reconstruction performance over m-TransH, but still struggles to capture fine-grained \(n\)-ary semantics and logical structure due to the absence of explicit positional or role modelling \cite{NaLP2021,S2S2021,ReAIE2023}.

% #######################################################################
\subsubsection{Tensor factorisation-based Models}

Tensor factorisation-based aware-less methods represent \(n\)-ary relational facts as entries of a higher-order tensor and factorise this tensor into structured low-rank components, without introducing explicit positional or role parameters. 

A straightforward strategy in this family is to extend existing binary KGRL models to the $n$-ary setting. Two early representatives are \textbf{r-SimplE} \cite{fatemi2019} and \textbf{m-DistMult}  \cite{fatemi2019}, which adapt SimplE and DistMult, respectively, by reusing their tensor factorisation structures.
In \textbf{r-SimplE} a higher-order fact \(r(e_1, e_2, e_3)\) is reified into binary triples \(r_1(e',e_1)\), \(r_2(e',e_2)\) and \(r_3(e',e_3)\), where \(e'\) is an auxiliary entity. 
This makes it easy to apply standard SimplE embeddings, but the auxiliary entities are sparse and lack rich neighbourhood context, which leads to low-quality embeddings and limited generalisation \cite{fatemi2019}.
In contrast, \textbf{m-DistMult} does not reify higher-order facts but generalises DistMult \cite{DistMult2014} to $n$-ary facts by extending its bilinear scoring function via element-wise multiplication:
\begin{equation}
\phi\big(r(e_i,\dots,e_j)\big) = \odot(r, e_i, \dots, e_j),
\label{eq:mdistmult}
\end{equation}
where \(\odot\) denotes the Hadamard product. 
This yields a very simple and efficient aware-less baseline, yet it inherits the main limitation of DistMult: it can only model symmetric relations and therefore has reduced expressiveness.

Moving beyond such direct extensions of binary models, \textbf{GETD} \cite{GETD2020} is the first tensor factorisation-based method designed specifically for $n$-ary facts.
It extends TuckER to an $n$-TuckER formulation and then applies tensor ring (TR) decomposition to factorise the higher-order core tensor into a circular sequence of third-order latent tensors. 
GETD offers strong expressiveness with linear complexity, but it only supports fixed-arity facts and is sensitive to the chosen decomposition settings \cite{NaLP2021,HyperMLN2022,stare2020,GRAN2021,HyConvE2023}.

\textbf{S2S} \cite{S2S2021} is proposed as an improvement over GETD to alleviate sparsity and over-parameterisation in the higher-order core tensor.
It partitions entity and relation embeddings into \(M\) segments and uses three diagonal operators to construct a sparse core tensor, which yields a complexity of \(O(m^{(n+1)})\) with \(m,n \ll d_{\text{feature}}\). 
This design reduces parameter growth and improves efficiency compared with GETD, but S2S still assumes fixed arity, remains aware-less in our taxonomy, and relies on careful choices of factorisation hyperparameters.

% Taken together, these tensor factorisationbased aware-less models move from direct extensions of binary KGRL via reification or symmetric multilinear products to dedicated \(n\)-ary tensor decompositions that improve expressiveness and efficiency. 
% \rev{At the same time, they share several limitations when modelling \(n\)-ary relational facts: most of them only support fixed-arity inputs, they do not capture asymmetric patterns or explicit role semantics, and they tend to be sensitive to factorisation strategies and decomposition configurations, which can harm generalisability across datasets \cite{NaLP2021,HyperMLN2022,stare2020,GRAN2021,HyConvE2023}.}

% #######################################################################
\subsubsection{Deep Neural Network-based Methods}

Deep neural network-based models learn \(n\)-ary relational representations through trainable architectures such as FCNs, CNNs, Transformer, GNNs, and HNNs.

% \paragraph{FCN-based models}

% Inspired by BoxE, \textbf{ShrinkE} \cite{ShrinkE2023} represents hyper-relational facts by explicitly modelling logical properties of qualifiers (monotonicity, implication, mutual exclusion). Entities are represented as points, and relations as spatial functions that map these points to query boxes via translation and rotation operations. Qualifiers will further "shrink" these boxes. Despite ShrinkE is lightweight, its qualifier monotonicity assumption, adding any qualifiers will not expand the set of possible answers, limits its effectiveness for opaque qualifiers \cite{ShrinkE2023}.

\paragraph{CNN-based models}

While previous CNN-based models such as HyConvE or HySAE strengthen relational expressiveness by tailoring convolutional filters to each entity position, they often lead to over-complicated architectures with high computational and memory costs. Addressing this trade-off between expressiveness and efficiency, \textbf{HyCubE} \cite{2025hycube} introduces a lightweight end-to-end architecture that combines 3D convolution with circular padding to capture full interactions among entities and relations. To generalise across both fixed- and mixed-arity tuples, HyCubE designs a novel alternate stacking and masking strategy, enabling position-agnostic yet structurally rich modelling without explicit position or role encodings. 
HyCubE thus achieves strong results while being 6 times faster, using half the memory, and 80\% fewer parameters than previous CNN-based models. 
However, its implicit encoding of roles and positions may limit fine-grained semantic modelling. Additionally, although lightweight among CNN-based methods, the 3D convolutions still entail non-negligible cost for large-scale industrial KHGs.

\paragraph{HNN-based models}
The HNN-based aware-less models discussed in section \ref{sec:ordered} and \ref{sec:roles} mainly target generic KHG or HKG completion. In contrast, several recent works apply HNNs to concrete application domains, showing how KHGs can support real-world reasoning tasks.

\textbf{Zhang et al.} \cite{zhang2024} study a domain-specific KHG built for bridge crane fault diagnosis. Their motivation is that traditional fault knowledge bases either rely only on text, or use binary KGs that cannot capture the $n$-ary relations among faults, components and symptoms. They construct an $n$-ary fault KHG and combine BERT-based text encoders with a hypergraph convolutional network to fuse semantic and structural features. A weighted similarity measure then retrieves similar faults for diagnosis support. 
This model demonstrates the practical value of HNNs on KHGs in industrial settings, but it is tied to a specific domain, depends on a manually curated ontology and treats entities within each hyperedge symmetrically, without explicit role or position modelling.

\textbf{KHG-Aclair} \cite{khg-aclair2024} targets recommender systems like SDK, but follows a simpler path. While SDK adopts a role-aware, dynamic HNN-based framework with self-supervised contrastive learning on user–item interactions, KHG-Aclair operates on a static item KHG and does not encode role or temporal semantics. This design choice aligns with its goal: to enhance long-tail item recommendations by leveraging high-order semantic relations from knowledge bases. To achieve this, it builds a hypergraph from Freebase-like resources and applies a dual hypergraph attention mechanism to aggregate relational contexts. A contrastive loss (inspired by KGCL \cite{KGCL2022k}) and PCA-based dimensionality reduction further boost embedding robustness. 
However, it is relatively rigid in design, as the attention is only used for neighbourhood aggregation and not to weigh the importance of entities within each $n$-ary fact. In spite of this, its streamlined architecture suits industrial use cases that require efficiency over expressivity.

\textbf{HyperQuery} \cite{2025hyperquery} takes a more general perspective and proposes a self-supervised HNN framework for both hyperedge prediction and KHG completion. Since many existing methods for higher-order LP focus on local structure and lack global semantics, HyperQuery first performs multilevel clustering to obtain global structural labels as pseudo-features, and then applies a novel edge-to-edge convolution scheme to aggregate information across hyperedges. This yields inductive node and hyperedge embeddings without requiring manual features. 
Nevertheless, HyperQuery introduces additional complexity through the clustering and convolution stages, and entities in each $n$-ary fact are still treated in a largely role- and position-agnostic way.

Overall, these three models illustrate how HNN-based KHG methods can be adapted to concrete domains such as industrial diagnosis \cite{zhang2024}, recommendation \cite{khg-aclair2024} and biomedical applications \cite{2025hyperquery}. At the same time, they highlight an open direction for future work: bringing explicit role and position semantics into practical, application-oriented HNN architectures on large-scale real-world hypergraphs.

% ####################################################################### 
% #######################################################################
\subsection{Temporal Extension}
\label{sec:temporal_ext}

Recent developments in temporal knowledge representation have extended static \(n\)-ary fact modelling to time-evolving or dynamic modelling. Temporal extensions to HTKGs (see Def. \ref{def:h-tkg}) and $n$-TKGs (see Def. \ref{def:$n$-tkg}) aim to capture temporal dependencies and entity dynamics over time. While some models \cite{hyperTKG2023} integrate time into the primary quadruples in hyper-relational facts, others \cite{NEnet2023} treat temporality via a set of timestamped graph snapshots.

\textbf{HypeTKG} \cite{hyperTKG2023} extends static hyper-relational modelling by introducing a temporal quadruple \((h, r, t, T)\), where \(T\) is a timestamp. It employs a Qualifier-Attentional Time-Aware Graph Encoder (QATGE) that aggregates entity representations from temporal neighbours through attention mechanisms. A Transformer-based Qualifier Matching Decoder (QMD) is then used to predict missing entities. Currently, HypeTKG only predicts missing entities from the main quadruples and does not support predictions in qualifiers or timestamps \cite{hyperTKG2023}.
% Moreover, the model explores the complementary role of time-invariant (TI) relational knowledge.

\textbf{NE-Net} \cite{NEnet2023} represents temporal facts as sequences of timestamped KGs, modelling each fact as \((predicate, \rho_1:e_1, \dots, \rho_n:e_n, t)\). NE-Net jointly learns entity and relation evolutions using two encoders. A CompGCN-based \cite{CompGCN2019} encoder models the overall entity-relation interactions from all concurrent facts at each timestamp, and an RGCN-based \cite{rgcn2018} encoder captures the interactions between core entities and relations. A transformer-based decoder is finally used to predict missing components in relational reasoning and entity reasoning under future timestamps.

Despite recent progress, temporal extensions for \(n\)-ary relational models remain under-explored. Future work may focus on predicting timestamps and qualifier information, incorporating fine-grained temporal dynamics, and improving scalability for real-world temporal \(n\)-ary facts.
% Both models are role-aware. HyperTKG encodes time explicitly, while Ne-Net models temporal dynamics through snapshot sequences.

\subsection{Complexity Analysis}

To make the above discussion more concrete, we summarise in Table~\ref{tab:complexity} the time and space complexity (per training step) of those $n$-ary relational models whose original papers explicitly provide complexity analyses.
Here, $n_e$ and $n_r$ denote the numbers of entities and relations; $n_r^{\mathrm{pri}}$ is the number of primary relations; $N$ is the number of $n$-ary facts (statements); 
$T$ is the number of timestamps; $d$ is the embedding dimensionality of entities and relations; $k$ is the latent space size for role embeddings; 
$m_a$ is the maximum arity of relations; $n_a$ is the maximum number of attribute-value qualifiers per fact; 
$\tilde{\Delta} = \max_i \deg(v_i)$ is the maximum node degree; $\Delta = \max_i \deg(e_i)$ is the maximum hyperedge degree; 
$L$ is the number of message-passing layers, $C$ is the number of conditional MP layers, $c$ is the number of TR latent tensors, and $e_r$ is the number of edges in the relation graph used by HYPER.

Overall, Table~\ref{tab:complexity} highlights a clear contrast between technical families.
Translation-based and most tensor factorisation-based models typically have linear or low-order polynomial time in $d$ and memory linear in $n_e$ and $n_r$, making them relatively lightweight.
FCN- and CNN-based models generally incur an $O(d^2)$ time cost but remain linear in the graph size.
In contrast, GNN- and HNN-based approaches, as well as hyperedge expansion methods, tend to have higher asymptotic cost, with complexity depending not only on $d$ but also on the number of facts $N$, timestamps $T$, and graph degrees ($\tilde{\Delta}$, $\Delta$).
This family-level comparison helps to explain which models are more suitable for large-scale deployments and which ones trade efficiency for greater expressive power.

\begin{table*}[t]
\centering
\caption{Time and space complexity of representative $n$-ary relational models.}
\label{tab:complexity}
\begin{tabular}{l l c c}
\hline
\textbf{Model} & \textbf{Technical family} & $\boldsymbol{O_{\text{time}}}$ & $\boldsymbol{O_{\text{space}}}$ \\
\hline
m-TransH \cite{mtransH2016} & Translation-based 
  & $O(d)$ 
  & $O(n_e d + 2 n_r d)$ \\

BoxE \cite{boxe2020} & Translation-based 
  & $O(m_a d)$ 
  & $O(n_e d + 2 n_r d m_a)$ \\

RAE \cite{RAE2018} & Translation-based 
  & $O(d^2)$ 
  & $O(n_e d + n_r d)$ \\

RAM \cite{RAM2021} & Tensor factorisation-based 
  & $O(d)$ 
  & $O(n_e d + k n_r d)$ \\

PosKHG \cite{PosKHG2023} & Tensor factorisation-based 
  & $O(d)$ 
  & $O(n_e d + k n_r m_a)$ \\

HSimplE \cite{fatemi2019} & Tensor factorisation-based 
  & $O(d)$ 
  & $O(n_e d + n_r d)$ \\

HypE \cite{fatemi2019} & Tensor factorisation-based 
  & $O(d)$ 
  & $O(n_e d + n_r d)$ \\

m-DistMult \cite{fatemi2019} & Tensor factorisation-based 
  & $O(d)$ 
  & $O(n_e d + n_r^{\mathrm{pri}} d)$ \\

GETD \cite{GETD2020} & Tensor factorisation-based 
  & $O(d^3)$ 
  & $O(n_e d + n_r d + c d^3)$ \\

S2S \cite{S2S2021} & Tensor factorisation-based 
  & $O(d)$ 
  & $O(n_e d + n_r d)$ \\

GAHE \cite{2025gahe} & Tensor factorisation-based 
  & $O(m_a d)$ 
  & $O(n_e d + n_r d m_a)$ \\

NaLP \cite{NaLP2021} & FCN-based 
  & $O(d^2)$ 
  & $O(n_e d + n_r d m_a)$ \\

NeuInfer \cite{neuinfer2020} & FCN-based 
  & $O(d^2)$ 
  & $O(n_e d + n_r d m_a)$ \\

HINGE \cite{HINGE2020} & CNN-based 
  & $O(d^2)$ 
  & $O(n_e d + n_r d m_a)$ \\

STARE \cite{stare2020} & GNN-based 
  & $O(N d^2 + n_a d^2)$ 
  & $O(n_e d + n_r d)$ \\

HypeTKG \cite{hyperTKG2023} & GNN-based 
  & $O(T n_e + T n_r)$ 
  & $O(n_e d + n_r d)$ \\

G-MPNN \cite{GMPNN2020} & HNN-based 
  & $O(N d^2)$ 
  & $O(n_e d + n_r^{\mathrm{pri}} d + n_a d)$ \\

HYPER \cite{2025hyper} & HNN-based 
  & $O(L e_r^{2} m_a^{2} d + n_r d^{2} + C(m_a n_r d + n_e d^{2} + N d^{2}))$ 
  & $O(n_e)$ \\

HyperQuery \cite{2025hyperquery} & HNN-based 
  & $O(n_r \tilde{\Delta} + n_e \Delta)$ 
  & \text{N/A} \\

TransEQ \cite{TransEQ2024} & Hyperedge expansion-based 
  & $O(N d^2)$ 
  & $O(n_e d + n_r d + N d)$ \\
\hline
\end{tabular}

\end{table*}

% ####################################################################### 
% #######################################################################
\subsection{Guidelines for $n$-ary Representation Models}

Designing effective representation learning models for \(n\)-ary relational data involves multiple dimensions. Based on our two-dimensional analysis and model taxonomy, we first summarise the desirable properties of a high-quality \(n\)-ary representation learning model as follows:
% of complexity, semantics, and scalability

\begin{itemize}
    \item \textbf{Position and Role Awareness.} The model should consider both entity position and semantic roles in \(n\)-ary facts. It should be able to capture entity-role correlations and the compatibility between entities, roles, and relations \cite{RAM2021, NaLP2021,HINGE2020}.
    
    \item \textbf{Full Expressiveness.} A good model should represent a wide range of relational types, including symmetric, antisymmetric, inverse, one-to-many, many-to-one, and many-to-many relational patterns \cite{GETD2020, fatemi2019, boxe2020, ReAIE2023,TransEQ2024,PosKHG2023}.
    
    \item \textbf{Relational Algebra Support.} Models should support basic relational algebra operations (e.g., projection, selection, union, renaming) that allow high-level abstraction and structured reasoning in KHGs. A fully expressive model does not mean that it can support relational algebra operations \cite{ReAIE2023}.
    
    \item \textbf{Support for Mixed-Arity Facts.} The model should be robust to datasets with fixed and mixed arity and adaptable to different relational formats \cite{S2S2021}.
    
    \item \textbf{Efficiency and Scalability.} Models need to scale to large knowledge bases with near-linear time and space complexity \cite{HyTransformer2021}.
    
    \item \textbf{Interpretability.} The reasoning results should be interpretable, ideally through transparent mechanisms such as logic-rule guided models (e.g., HyperMLN), although such mechanisms are underexplored in the KHG setting \cite{HyperMLN2022}.
    
    \item \textbf{Multimodal Capability.} The model should handle different input types including categorical entities, numerical data, textual descriptions, and temporal information \cite{HyNT2023}.
\end{itemize}

In summary, a high-quality \(n\)-ary relational representation learning model should strike a balance between \textit{semantic expressiveness}, \textit{computational efficiency}, and \textit{interpretability}, while being generalisable to dynamic, multimodal, and structurally heterogeneous knowledge sources.

% #######################################################################
\section{\rev{Benchmark and Training Settings}}
\label{sec:ds}

\rev{This section summarises the benchmark resources and training settings used to evaluate $n$-ary relational representation learning models. 
We first review representative tuple-based and qualifier-based datasets for KHG and HKG modelling, respectively. 
We then discuss negative sampling strategies, which are essential for training link prediction models. 
Finally, we provide benchmark-style syntheses of reported KHG and HKG model performances. 
Since existing studies differ in dataset preprocessing, splits, prediction targets, sampling strategies, and evaluation protocols, these results should be interpreted as cross-paper summaries rather than strictly unified comparisons.}

% In this section, we present the benchmark datasets that are commonly used to evaluate \(n\)-ary relational representation learning models, including both KHG and HKG structures. In addition to the dataset statistics and structure, we also discuss negative sampling strategies, which play a crucial role in the training of link prediction models. Although negative sampling is a training technique, we include it here to provide a comprehensive view of how benchmark datasets are used and extended during model learning.

% #######################################################################
\subsection{Benchmark Datasets}

\rev{
Most existing KGs represent facts in the triple format $(s, r, o)$, following the RDF standard \cite{RDF2015}. 
However, many real-world facts involve more than two entities and roles. 
Studies show that in Freebase \cite{Freebase2008}, 61\% of relations are non-binary \cite{fatemi2019}, and over one-third of entities participate in $n$-ary relations \cite{mtransH2016}. Similar phenomena are also observed in Wikidata \cite{wikidata2014} and YAGO \cite{yago2007}. 
Nevertheless, several classical KG benchmarks, such as FB15K \cite{FB15K2013}, were constructed by transforming $n$-ary facts into triples, which may lead to irreversible information loss \cite{mtransH2016}.

These observations motivate the need for dedicated benchmarks that preserve $n$-ary relational structures. 
However, due to data acquisition difficulties, heterogeneous data formats, and model complexity, standard benchmarks for $n$-ary relational representation learning remain relatively limited \cite{ReAIE2023}. 
We organise existing datasets into two main categories: tuple-based datasets and qualifier-based datasets. }
Detailed dataset statistics and individual descriptions, including construction details, limitations, and dataset-specific properties, are provided in Supplementary Material, Section~\Romannum{2}.

\paragraph{Tuple-based datasets}
Tuple-based datasets represent facts as $n$-ary tuples, where $n \geq 2$, usually written as $r(e_1,e_2,\dots,e_n)$. 
They are commonly used to evaluate KHG-style models that directly encode higher-order relational facts. 
Representative examples include JF17K, FB-AUTO, m-FB15K, and their fixed-arity or temporal variants.

\paragraph{Qualifier-based datasets}
Qualifier-based datasets represent $n$-ary facts as primary triples enriched with qualifier pairs. 
They explicitly distinguish the main relational statement from auxiliary contextual information and are commonly used to evaluate HKG-style models. 
Representative examples include WikiPeople, JF17K, WD50K, numerical or temporal variants, and recommendation-oriented datasets.

% #######################################################################
\subsection{Negative Sampling in KHGs}

\rev{
Negative sampling plays an important role in training link prediction models, but most existing KG sampling strategies are designed for binary triples and cannot be directly transferred to $n$-ary relational data. 
For example, methods such as NSCaching \cite{nscaching2019} require position-specific caches, SANS \cite{ahrabian2020} assumes a meaningful $k$-hop neighbourhood structure, and GAN-based methods often introduce additional training cost \cite{kbgan2017,ksgan2019,igan2018}. 
By contrast, self-adversarial sampling \cite{rotate2019}, which selects negatives according to model scores, remains more flexible and can be adapted to KHG settings.

In practice, most KHG models still rely on uniform random sampling \cite{hypergan2022}. 
Given a positive $n$-ary fact, negative samples are generated by randomly replacing one entity at a time at different argument positions. 
Although simple and efficient, this strategy often produces low-quality negatives with mismatched entity types or irrelevant semantics, leading to the zero-loss problem and reducing training efficiency \cite{igan2018}. 
To address this issue, HyperGAN \cite{hypergan2022} introduces a GAN-based negative sampling strategy for KHGs, where a generator produces more informative negative samples and a discriminator distinguishes them from true facts. }
A formal description of uniform negative sampling for KHGs is provided in Supplementary Material, Section~\Romannum{3}.
More generally, designing type-aware, structure-aware, and efficient negative sampling strategies remains an important direction for improving the robustness of $n$-ary relational representation learning models.

\subsection{\rev{Benchmark Model Performances}}

\rev{
To provide a more systematic empirical view of existing $n$-ary relational representation learning methods, we summarise reported benchmark results for representative KHG and HKG models. 
Rather than presenting a newly rerun unified benchmark, this subsection provides a benchmark-style synthesis based on results reported in the original papers. 
Since existing studies differ in dataset preprocessing, train/validation/test splits, negative sampling strategies, filtered evaluation protocols, hyperparameter settings, and prediction targets, the results should be interpreted as indicative empirical trends rather than fully comparable scores under a single experimental protocol. 
Following the two major data structures discussed in this survey, we organise the results into KHG and HKG benchmarks, reporting each model's technical family, semantic-awareness category, and standard ranking metrics such as MRR, Hits@1, Hits@3, and Hits@10 when available.
}

\subsubsection{\rev{KHG Benchmark Results}}

\rev{
Table~\ref{tab:khg_benchmark_synthesis} summarises reported link prediction results of representative KHG models on three commonly used mixed-arity benchmarks: JF17K, WikiPeople, and FB-AUTO \cite{2025hysae,ReAIE2023}.
For each model, we report its technical family, semantic-awareness category, and standard ranking metrics, including MRR, Hits@1, Hits@3, and Hits@10. 
The ``--'' symbol indicates unreported results, while $^\dagger$ denotes results that are locally reproduced or reported as such in the corresponding source. 
Additional results on fixed-arity KHG subsets, such as JF17K-3/4 and WikiPeople-3/4, are provided in} Supplementary Material, Section~\Romannum{4}.A.

\rev{
The results show that recent position-aware models, such as HyConvE, HJE, and HySAE, generally achieve strong performance on mixed-arity KHG benchmarks. 
This suggests that preserving the structural order of entities is particularly important for tuple-based KHGs, where the position of an entity often determines its semantic contribution to the fact. 
Role-aware modelling remains valuable, but current role-aware KHG models do not always outperform recent position-aware neural architectures. 
This may be partly because roles in KHG datasets are often implicit or derived from relation-specific positions, while some role-aware models rely on relatively linear tensor formulations or introduce additional role-specific parameters. 
These observations suggest that future KHG models should jointly model positional structure, role semantics, and high-order entity interactions in efficient architectures, while being evaluated under more standardised benchmark settings.
}

\begin{table*}[t]
\centering
\caption{\rev{Cross-paper synthesis of reported link prediction results on commonly used mixed-arity KHG benchmarks.}}
\label{tab:khg_benchmark_synthesis}
\tiny
\setlength{\tabcolsep}{1.6pt}
\renewcommand{\arraystretch}{1.05}

\resizebox{\textwidth}{!}{
\begin{tabular}{@{}lllcccccccccccc@{}}
\hline
\textbf{Model} 
& \textbf{Family} 
& \textbf{Awareness}
& \multicolumn{4}{c}{\textbf{JF17K}}
& \multicolumn{4}{c}{\textbf{WikiPeople}}
& \multicolumn{4}{c}{\textbf{FB-AUTO}} \\
\cline{4-15}
& & 
& \textbf{MRR} & \textbf{H@1} & \textbf{H@3} & \textbf{H@10}
& \textbf{MRR} & \textbf{H@1} & \textbf{H@3} & \textbf{H@10}
& \textbf{MRR} & \textbf{H@1} & \textbf{H@3} & \textbf{H@10} \\
\hline

m-TransH
& Translation-based
& Position-aware
& 0.444 & 0.370 & 0.475 & 0.581
& -- & -- & -- & --
& 0.728 & 0.727 & 0.728 & 0.728 \\

% BoxE
% & Translation-based
% & Position-aware
% & 0.560 & 0.472 & 0.606 & 0.722
% & 0.395 & 0.293 & 0.453 & 0.503
% & 0.844 & 0.814 & 0.863 & 0.898 \\

RAE       
& Translation-based 
& Aware-less   
& 0.392 & 0.312 & 0.433 & 0.561 
& 0.253 & 0.118 & 0.343 & 0.463 
& 0.703 & 0.614 & 0.764 & 0.854 \\

m-DistMult
& Tensor factorisation-based
& Aware-less
& 0.463 & 0.372 & 0.510 & 0.634
& -- & -- & -- & --
& 0.784 & 0.745 & 0.815 & 0.845 \\

m-CP
& Tensor factorisation-based
& Position-aware
& 0.392 & 0.303 & 0.441 & 0.560
& -- & -- & -- & --
& 0.752 & 0.704 & 0.785 & 0.837 \\

% NaLP      
% & FCN-based 
% & Role-aware 
% & 0.310 & 0.239 & 0.334 & 0.450 
% & 0.338 & 0.272 & 0.362 & 0.466 
% & 0.672 & 0.611 & 0.712 & 0.774 \\

% HINGE     
% & CNN-based 
% & Role-aware 
% & 0.473 & 0.397 & 0.490 & 0.618 
% & 0.333 & 0.259 & 0.361 & 0.477 
% & 0.678 & 0.630 & 0.772 & 0.774 \\

% NeuInfer  
% & FCN-based 
% & Role-aware 
% & 0.451 & 0.373 & 0.484 & 0.604 
% & 0.350 & 0.282 & 0.381 & 0.467 
% & 0.737 & 0.700 & 0.755 & 0.805 \\

HypE      
& Tensor factorisation-based
& Position-aware 
& 0.494 & 0.399 & 0.532 & 0.650 
& 0.263 & 0.127 & 0.355 & 0.486 
& 0.804 & 0.774 & 0.824 & 0.856 \\

HSimplE   
& Tensor factorisation-based 
& Position-aware  
& 0.472 & 0.408 & 0.538 & 0.656
& 0.227$^\dagger$ & 0.221$^\dagger$ & 0.298$^\dagger$ & 0.351$^\dagger$ 
& 0.798 & 0.766 & 0.821 & 0.855 \\

S2S       
& Tensor factorisation-based 
& Aware-less   
& 0.528 & 0.457 & 0.570 & 0.690 
& 0.372 & 0.277 & 0.439 & 0.533 
& -- & -- & -- & -- \\

% tNaLP+    
% & CNN-based 
% & Role-aware 
% & 0.449 & 0.370 & 0.484 & 0.598 
% & 0.339 & 0.269 & 0.369 & 0.473 
% & 0.729 & 0.645 & 0.748 & 0.826 \\

RAM       
& Tensor factorisation-based 
& Role-aware 
& 0.539 & 0.463 & 0.573 & 0.690 
& 0.380 & 0.279 & 0.445 & 0.539 
& 0.830 & 0.803 & 0.851 & 0.876 \\

HyperMLN  
& Logic rule-based 
& Role-aware 
& 0.556 & 0.482 & 0.597 & 0.717 
& -- & -- & -- & -- 
& 0.831 & 0.803 & 0.851 & 0.877 \\

PosKHG    
& Tensor factorisation-based 
& Role-aware 
& 0.545 & 0.469 & 0.582 & 0.706 
& 0.315$^\dagger$ & 0.214$^\dagger$ & 0.377$^\dagger$ & 0.475$^\dagger$ 
& 0.856 & 0.821 & 0.876 & 0.895 \\

ReAIE     
& Tensor factorisation-based 
& Position-aware  
& 0.559 & 0.482 & 0.594 & 0.705 
& 0.332$^\dagger$ & 0.207$^\dagger$ & 0.417$^\dagger$ & 0.514$^\dagger$ 
& 0.873 & 0.852 & 0.886 & 0.909 \\

% RD-MPNN   
% & GNN-based 
% & Position-aware  
% & 0.512 & 0.445 & 0.573 & 0.685 
% & -- & -- & -- & -- 
% & 0.810 & 0.714 & 0.880 & 0.888 \\

HyConvE   
& CNN-based 
& Position-aware  
& 0.580 & 0.478 & 0.610 & 0.729 
& 0.362 & 0.275 & 0.388 & 0.501 
& 0.847 & 0.820 & 0.872 & 0.901 \\

HJE       
& CNN-based 
& Position-aware  
& 0.590 & 0.507 & 0.613 & 0.729 
& 0.450 & 0.375 & 0.487 & 0.582 
& 0.872 & 0.848 & 0.886 & 0.903 \\

HyCubE    
& CNN-based 
& Aware-less   
& 0.584 & 0.508 & 0.616 & 0.730 
& 0.448 & 0.368 & 0.490 & 0.592 
& 0.881 & 0.860 & 0.894 & 0.918 \\

HyCubE+   
& CNN-based 
& Aware-less   
& 0.582 & 0.511 & 0.611 & 0.720 
& 0.433 & 0.347 & 0.478 & 0.591 
& 0.891 & 0.872 & 0.901 & 0.923 \\

HySAE-I   
& CNN-based 
& Position-aware  
& \textbf{0.596} & \textbf{0.521} & \textbf{0.628} & \textbf{0.742 }
& \textbf{0.454} & 0.372 & \textbf{0.496} & 0.601 
& 0.892 & 0.873 & 0.902 & \textbf{0.926} \\

HySAE-E   
& CNN-based 
& Position-aware  
& 0.592 & 0.516 & 0.626 & 0.741 
& \textbf{0.454} & \textbf{0.373} & 0.495 & \textbf{0.603 }
& \textbf{0.893} & \textbf{0.876} & \textbf{0.904} & 0.924 \\

\hline
\end{tabular}
}

\vspace{0.5ex}
\begin{minipage}{\textwidth}
\footnotesize
\textit{Note}: ``--'' indicates that the result is not reported in the corresponding source. 
Results are grouped as a cross-paper synthesis rather than a strictly unified reproduction, since different papers may adopt different splits, preprocessing, rerun settings, and hyperparameter choices. 
$^\dagger$ denotes results locally reproduced or reported as such in the corresponding source.
When multiple reported values exist for the same model, we retain the value from the most recent benchmark table using the same dataset collection whenever possible.
\end{minipage}

\end{table*}

\subsubsection{\rev{HKG Benchmark Results}}

\rev{
Table~\ref{tab:hkg_benchmark_synthesis} reports a benchmark-style synthesis of representative HKG models on three commonly used benchmarks: JF17K, WikiPeople, and WD50K \cite{2024hypercl, HAHE2023}. 
The table focuses on all-entity prediction, where the missing entity in a hyper-relational fact $(s, r, o, \{(a_i, v_i)\}_{i=1}^{n})$ may appear either in the primary triple, i.e., $s$ or $o$, or in qualifier value positions, i.e., $v_1,\ldots,v_n$, when supported by the model. 
For each model, we report its technical family, semantic-awareness category, and the most consistently reported metrics, namely MRR, Hits@1, and Hits@10.
}

Additional HKG completion results on WikiPeople, JF17K, and FB-AUTO are provided in Supplementary Material, Section~\Romannum{4}.B.
\rev{
% These results are not merged with Table~\ref{tab:hkg_benchmark_synthesis} because they use a different dataset collection, replacing WD50K with FB-AUTO, and additionally report Hits@3. 
% They are therefore provided as a complementary reference for readers interested in this alternative benchmark setting.
These results are not merged with Table~\ref{tab:hkg_benchmark_synthesis}, because they come from an alternative HKG completion setting reported in prior studies~\cite{ShrinkE2023,hyperformer2023,HAHE2023}.
Unlike the main HKG table, which follows the all-entity prediction setting on JF17K, WikiPeople, and WD50K, this alternative setting replaces WD50K with FB-AUTO and additionally reports Hits@3.
Since the two settings may also differ in preprocessing choices, prediction targets, and rerun protocols, we provide these results separately as complementary references rather than merging them into a single table.
}

\rev{
Several observations can be drawn from Table~\ref{tab:hkg_benchmark_synthesis}. 
Most HKG models are role-aware under our taxonomy, as they explicitly encode qualifier relations, qualifier values, or attribute-value pairs. 
Thus, performance differences are mainly driven not by whether qualifier roles are represented, but by how models capture interactions among the primary triple, qualifiers, and surrounding graph structure. 
Early models such as NaLP, NeuInfer, and HINGE already incorporate qualifier information, while later neural models further improve hyper-relational modelling through graph message passing, Transformer-based interaction, heterogeneous attention, ontology information, or contrastive learning. 
In particular, HyperCL achieves the best reported performance across the three datasets by combining hierarchical ontology information with contrastive learning, suggesting that HKG completion benefits from both internal fact modelling and external semantic constraints for entity discrimination.
}

\subsubsection{\rev{Comparison between the KHG and HKG benchmark tables}}

\rev{
Comparing the KHG and HKG benchmark tables suggests that HKG models generally achieve strong performance on the common datasets JF17K and WikiPeople.
This tendency should be interpreted cautiously, as the two groups of models may rely on different preprocessing choices, prediction scopes, and reported settings.  
Nevertheless, it indicates that the qualifier-based HKG formalism can be effective when the primary triple and auxiliary information play different semantic roles. 
By preserving the distinction between the main relational statement and its qualifiers, HKG models can explicitly capture how additional attributes refine or condition the primary fact, whereas tuple-based KHG models encode all participating entities within a single ordered tuple and may require stronger role modelling to separate core and contextual information. 
}

\rev{
Future work should further investigate how primary triples and qualifiers interact at different semantic levels. 
Ontology-aware, contrastive, and structure-aware mechanisms are promising for improving entity discrimination, while qualifier-value/-relation prediction, and multi-position prediction deserve more attention because they better reflect the incompleteness of real hyper-relational facts. 
Finally, more standardised benchmark suites with unified dataset versions, prediction targets, and negative sampling strategies are still needed to distinguish genuine modelling improvements from differences in experimental settings.
}

\begin{table*}[t]
\centering
\caption{\rev{Cross-paper synthesis of reported all-entity prediction results on commonly used HKG benchmarks}.}
\label{tab:hkg_benchmark_synthesis}
\tiny
\setlength{\tabcolsep}{2.0pt}
\renewcommand{\arraystretch}{1.05}

\resizebox{\textwidth}{!}{
\begin{tabular}{@{}lllccccccccc@{}}
\hline
\textbf{Model} 
& \textbf{Family} 
& \textbf{Awareness}
& \multicolumn{3}{c}{\textbf{JF17K}}
& \multicolumn{3}{c}{\textbf{WikiPeople}}
& \multicolumn{3}{c}{\textbf{WD50K}} \\
\cline{4-12}
& & 
& \textbf{MRR} & \textbf{H@1} & \textbf{H@10}
& \textbf{MRR} & \textbf{H@1} & \textbf{H@10}
& \textbf{MRR} & \textbf{H@1} & \textbf{H@10} \\
\hline

NaLP
& FCN-based
& Role-aware
& 0.364 & 0.287 & 0.519
& 0.327 & 0.265 & 0.449
& 0.223 & 0.162 & 0.337 \\

tNaLP
& FCN-based
& Role-aware
& 0.370 & 0.292 & 0.528
& 0.333 & 0.272 & 0.457
& 0.230 & 0.169 & 0.344 \\

NeuInfer
& FCN-based
& Role-aware
& 0.489 & 0.418 & 0.625
& 0.349 & 0.281 & 0.506
& 0.235 & 0.178 & 0.355 \\

HINGE
& CNN-based
& Role-aware
& 0.519 & 0.445 & 0.682
& 0.367 & 0.305 & 0.488
& 0.245 & 0.181 & 0.362 \\

sHINGE
& CNN-based
& Role-aware
& 0.528 & 0.459 & 0.701
& 0.372 & 0.312 & 0.499
& 0.248 & 0.185 & 0.370 \\

ShrinkE
& FCN-based 
& Role-aware
& 0.554 & 0.459 & 0.702
& 0.472 & 0.415 & 0.589
& 0.336 & 0.259 & 0.478 \\

GRAN
& Transformer-based
& Role-aware
& 0.652 & 0.579 & 0.798
& 0.496 & 0.426 & 0.619
& 0.361 & 0.287 & 0.504 \\

% MSeaHKG
% & GNN-based
% & Role-aware
% & 0.579 & 0.481 & 0.718
% & 0.393 & 0.301 & 0.562
% & 0.324 & 0.239 & 0.481 \\

HyNT
& Transformer-based
& Role-aware
& 0.634 & 0.558 & 0.783
& 0.457 & 0.376 & 0.597
& 0.337 & 0.271 & 0.464 \\

HyperFormer
& GNN-based
& Role-aware
& 0.651 & 0.587 & 0.784
& 0.471 & 0.356 & 0.645
& 0.365 & 0.285 & 0.514 \\

StarE
& GNN-based
& Role-aware
& 0.584 & 0.504 & 0.741
& 0.394 & 0.290 & 0.593
& 0.315 & 0.240 & 0.458 \\

Hy-Transformer
& Transformer-based
& Role-aware
& 0.592 & 0.513 & 0.750
& 0.399 & 0.298 & 0.588
& 0.314 & 0.241 & 0.453 \\

QUAD
& GNN-based
& Role-aware
& 0.585 & 0.504 & 0.747
& 0.379 & 0.272 & 0.583
& 0.316 & 0.245 & 0.451 \\

HAHE
& GNN-based
& Role-aware
& 0.657 & 0.585 & 0.798
& 0.495 & 0.421 & 0.623
& 0.379 & 0.305 & 0.521 \\

% DHGE
% & Dual-view embedding
% & Role-aware
% & 0.556 & 0.467 & 0.718
% & 0.457 & 0.406 & 0.572
% & 0.305 & 0.231 & 0.501 \\

HyperCL+StarE
& Contrastive learning+GNN
& Role-aware
& 0.602 & 0.528 & 0.768
& 0.415 & 0.309 & 0.618
& 0.336 & 0.266 & 0.487 \\

HyperCL+Hy-Transformer
& Contrastive learning+GNN
& Role-aware
& 0.617 & 0.534 & 0.778
& 0.419 & 0.316 & 0.615
& 0.338 & 0.263 & 0.488 \\

HyperCL+QUAD
& Contrastive learning+GNN
& Role-aware
& 0.605 & 0.529 & 0.776
& 0.396 & 0.288 & 0.605
& 0.339 & 0.268 & 0.498 \\

HyperCL+HAHE
& Contrastive learning+GNN
& Role-aware
& \textbf{0.673} & \textbf{0.604} & \textbf{0.818}
& \textbf{0.509} & \textbf{0.437} & \textbf{0.644}
& \textbf{0.395} & \textbf{0.321} & \textbf{0.539} \\

\hline
\end{tabular}
}
\vspace{0.5ex}
\begin{minipage}{\textwidth}
\footnotesize
\textit{Note}: Results are mainly collected from the all-entity prediction setting reported in HyperCL, so that base HKG encoders and HyperCL-enhanced variants are compared under the same benchmark setting. 
Scores may differ from those reported in the original model papers due to different preprocessing, evaluation protocols, and re-run settings.
\end{minipage}

\end{table*}

% ####################################################################### 
\section{Summary \& Open Challenges}
\label{sec:conclusion}

\rev{
In this survey, we systematically investigated representation learning methods for $n$-ary relational knowledge bases. 
Prior surveys have provided valuable overviews of KG, HG, and NKG representation learning, but they provide limited analysis of semantic-awareness differences, qualifier modelling mechanisms, training settings, and cross-paper empirical trends.
To address these limitations, we proposed a two-dimensional taxonomy that jointly considers technical methodologies and semantic-awareness levels, i.e., aware-less, position-aware, and role-aware modelling. 
This dual-axis view clarifies how existing KHG and HKG models process $n$-ary facts, capture entity positions, semantic roles, and qualifier structures, and differ in their modelling assumptions and representational capabilities.

Beyond taxonomy, we unified the terminology and formalism of $n$-ary relational data, clarified the corresponding link prediction formulations, and summarised representative models, benchmark datasets, training settings, negative sampling strategies, and reported model performances. 
In particular, our benchmark-style syntheses of KHG and HKG results provide a structured empirical overview, while also highlighting the difficulty of fully unified comparison due to heterogeneous datasets, splits, prediction targets, and evaluation protocols. 
The main open challenges are summarised as follows:
}

\begin{itemize}
    \item \textbf{\rev{Unified benchmarks and evaluation protocols}:}
    Current comparisons remain affected by heterogeneous dataset versions, preprocessing choices, prediction targets, negative sampling strategies, and filtered evaluation protocols. 
    Standardised benchmark suites with clearly defined tasks, splits, metrics, and evaluation settings are therefore essential for reliable model comparison.

    \item \textbf{End-to-end and extensible architectures:} 
    Many $n$-ary models still rely on redundant feature mapping procedures, which increases parameter complexity and may weaken interactions among elements. 
    Recent end-to-end models offer promising solutions for compact mixed-arity modelling \cite{2025hycube,2025hysae}, but scalable architectures that generalise across KHGs, HKGs, and downstream tasks remain challenging.
    
    \item \textbf{High-quality negative sampling:} 
    Uniform sampling may generate low-quality negatives and suffer from the zero-loss problem \cite{igan2018}. 
    Future work could further explore adversarial \cite{rotate2019}, GAN-based \cite{hypergan2022}, or structure-aware sampling strategies to improve robustness and training efficiency.

    \item \textbf{Temporal extensions:} 
    Although this survey mainly focuses on static $n$-ary relational representation learning, temporal information is important in many real-world knowledge bases. 
    Recent works such as HypeTKG \cite{hyperTKG2023} and NE-Net \cite{NEnet2023} provide initial attempts, but systematic temporal $n$-ary modelling remains underexplored.

    \item \textbf{Inductive generalisation:} 
    Most existing methods assume a transductive setting where entities and relations are observed during training \cite{stare2020,mtransH2016,fatemi2019,NaLP2021,HINGE2020,GETD2020}. 
    Recent works such as MAYPL \cite{maypl2025} and HyperQuery \cite{2025hyperquery} have begun to study inductive prediction and query answering, but generalisation to unseen entities, relations, and facts remains an open problem.

    \item \textbf{Interpretability:} 
    Many $n$-ary representation learning models remain black-box systems, limiting their use in high-stakes domains. 
    Logic-guided approaches such as HyperMLN \cite{HyperMLN2022} are promising, but more interpretable frameworks are needed to expose reasoning paths, role interactions, and model decisions.

    \item \textbf{Multimodal information fusion:} 
    Real-world knowledge often includes textual, numerical, visual, and structural information. 
    Recent works such as HyNT \cite{HyNT2023} and HyperFM \cite{2025hyperfm} show the potential of numerical and multimodal signals, while lightweight and modality-robust $n$-ary modelling remains an open challenge.
\end{itemize}

Overall, $n$-ary relational representation learning is moving towards more structured taxonomies, richer semantic modelling, and more systematic empirical evaluation. 
Addressing these challenges will further improve the scalability, interpretability, and practical applicability of KHG and HKG representation learning methods.

\section{Acknowledgments}
\noindent This publication is based upon work from COST Action CA23147
GOBLIN - Global Network on Large-Scale, Cross-domain and Multilingual Open KGs, supported by COST (European Cooperation in Science and Technology, https://www.cost.eu).

\bibliographystyle{IEEEtran}
\bibliography{khg_survey}

\begin{IEEEbiography}[{\includegraphics
[width=1in,height=1.25in,clip,
keepaspectratio]{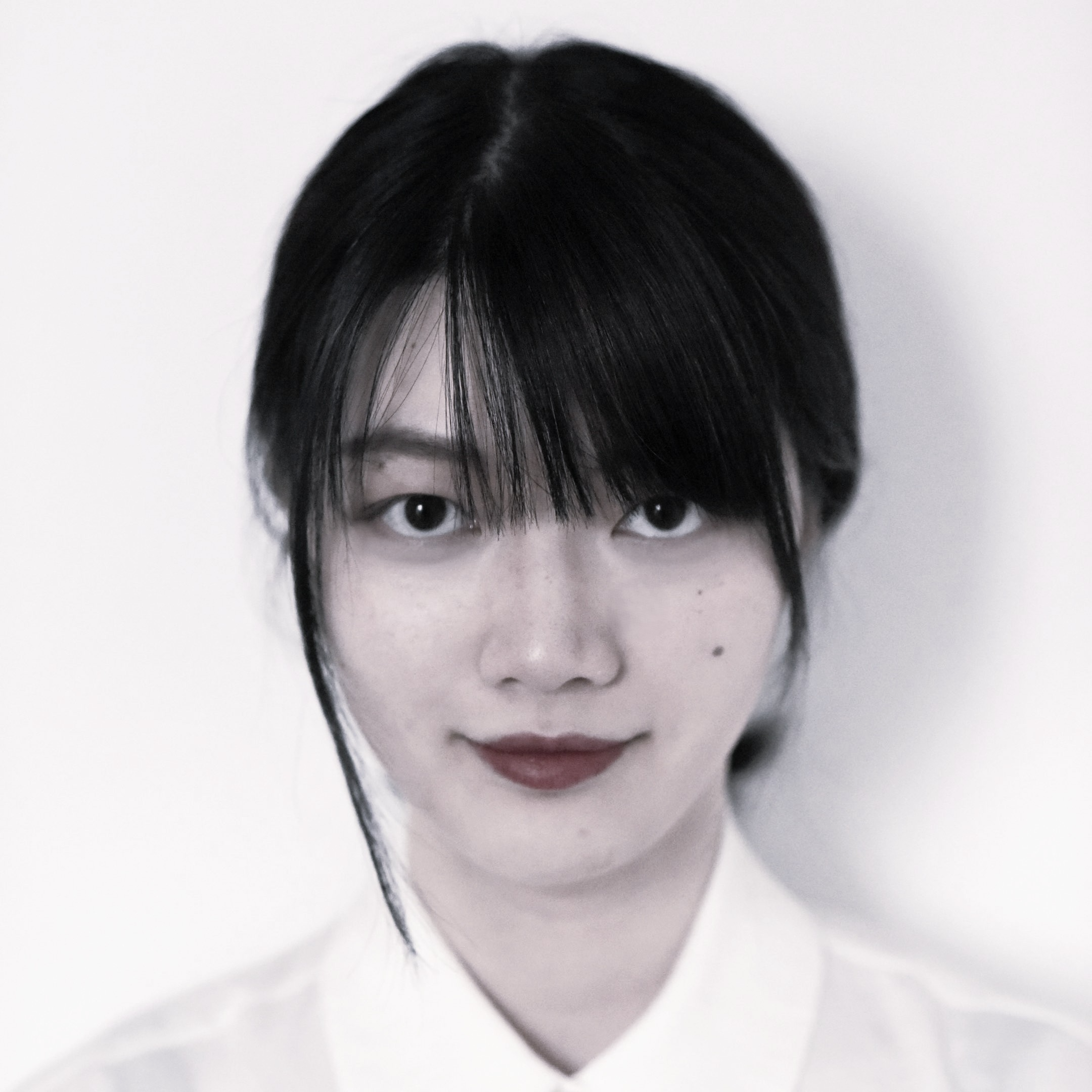}}]{Xiaohua Lu}
received the Engineering degree in Mathematical Modelling from INSA Rouen and the M.Sc. in Data Science and Engineering from the University of Rouen in France. She is currently collaborating on research in Graph Representation Learning. Her research interests include knowledge hypergraphs, temporal reasoning, and graph-based machine learning.
\end{IEEEbiography}

\begin{IEEEbiography}[{\includegraphics
[width=1in,height=1.25in,clip,
keepaspectratio]{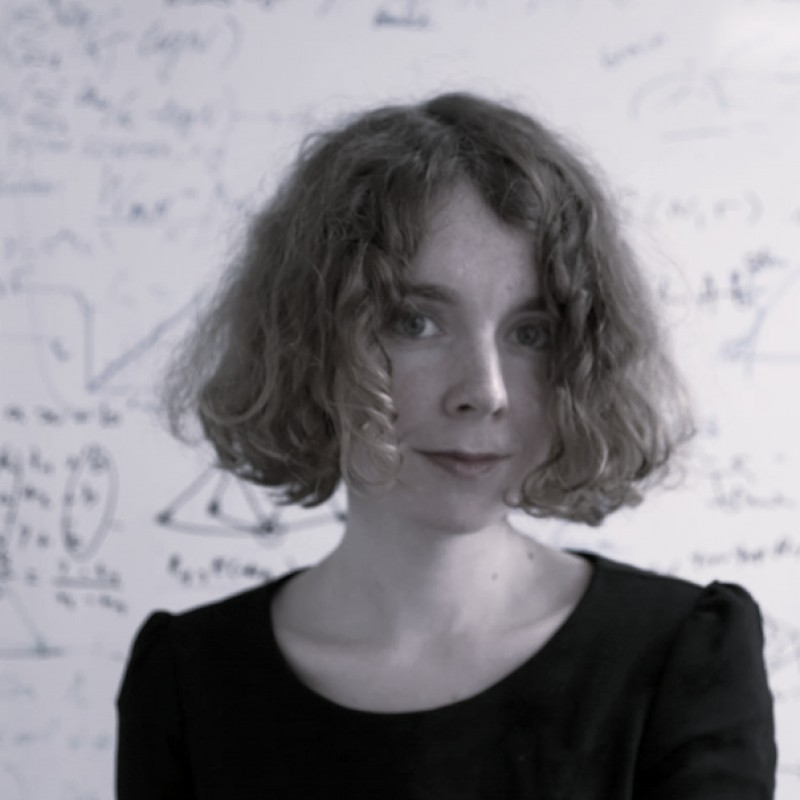}}]{Liubov Tupikina}
is a post-doctoral researcher in the group of Dr. Denis Grebenkov at the laboratory of Condensed Matter Physics of Ecole Polytechnique. She has graduated from Lomonosov Moscow State University, Russia (Diploma in Mathematics). She defended her PhD thesis in theoretical physics in Humboldt University of Berlin as part of Marie-Curie LINC project in Potsdam Climate Institute Germany in the group of Interdisciplinary methods on the topic “Temporal and spatial aspects of correlation networks and dynamical network models – analytical approaches and physical applications”. Her interests are in the theory of stochastic processes, statistical physics of complex systems, in particular, understanding the transport phenomena in complex systems and anomalous diffusion in living cells.
\end{IEEEbiography}

\begin{IEEEbiography}[{\includegraphics
[width=1in,height=1.25in,clip,
keepaspectratio]{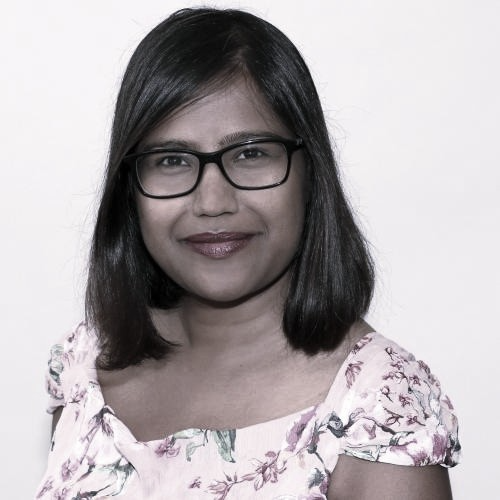}}]{Mehwish Alam }
is an Associate Professor and is leading a research group in Neurosymbolic Artificial Intelligence at Télécom Paris, Institut Polytechnique de Paris, France. Her research topics include Language Models, Natural Language Processing, Graph Learning, Machine/Deep Learning, and KGs. More specifically her research group focuses on the topics such as knowledge based hallucination detection in language models, learning representation from hyper and dynamic graphs, cognitive aspects of language models, explainability, and the application domains such as biomedicine and cultural heritages. Her research work has been published in high ranking conferences such as IJCAI, SIGIR, CIKM, ISWC, ECAI, and LREC. She has also published in prestigious journals such as Knowledge Based Systems, IEEE Computational Intelligence Magazine, Social Network Analysis and Mining, Scientometrics, Progress in Artificial Intelligence, and Semantic Web Journal. She is also part of the team leading the development of YAGO KG. She is currently leading academic and industrial collaborations on various aspects on the intersection of LLMs and KGs. She has been a part of the Organization Committee of various conferences in the field of Semantic Web and is a Senior PC member in ECAI and LREC/COLING. 
\end{IEEEbiography}

\vfill

\end{document}